%% file: neurips_2026.tex
\documentclass{article}

 \usepackage[main, final]{neurips_2026}

\usepackage[utf8]{inputenc} 
\usepackage[T1]{fontenc}    
\usepackage{hyperref}       
\usepackage{url}            
\usepackage{booktabs}       
\usepackage{amsfonts}       
\usepackage{nicefrac}       
\usepackage{microtype}      
\usepackage{xcolor}         

\usepackage{amsmath}

\usepackage{graphicx}
\usepackage{duckuments}

\usepackage{adjustbox} 

\usepackage{subcaption}
\usepackage{multirow}
\usepackage{wrapfig}

\newcommand{\paragrapht}[1]{\vspace{-10pt}\paragraph{#1}}
\usepackage{enumitem}

\usepackage{algorithm}
\usepackage{algpseudocode}
\usepackage{booktabs}
\usepackage{multirow}

\title{VideoMaMa++: Temporally Consistent Video Matting via Preserve-and-Refine Tokens}

\author{
    Sangbeom Lim\textsuperscript{1} \qquad
    Seoung Wug Oh\textsuperscript{2} \qquad
    Heeji Yoon\textsuperscript{3} \\
    \bfseries
    Seungryong Kim\textsuperscript{3\dag} \qquad
    Joon-Young Lee\textsuperscript{2\dag} \\[0.8em]
    \textsuperscript{1}Korea University \qquad
    \textsuperscript{2}Adobe Research \qquad
    \textsuperscript{3}KAIST AI\\[0.8em]
    {\tt \href{https://cvlab-kaist.github.io/VideoMaMa-PlusPlus}{{https://cvlab-kaist.github.io/VideoMaMa-PlusPlus }}}
}

\begin{document}

\maketitle
\begingroup
\renewcommand{\thefootnote}{\dag}
\footnotetext{Co-corresponding authors.}
\endgroup

\input{Figures/teaser}

\begin{abstract}
    Diffusion-based video matting has recently emerged as a leading approach, leveraging video diffusion priors from internet-scale pretraining to achieve strong zero-shot generalization from limited synthetic data. However, the heavy compute of these diffusion backbones limits inference to a short chunk of just 6--24 frames, so even ordinary videos span multiple chunks. Because each chunk is matted independently from coarse mask guidance, existing methods fail to maintain consistency across chunk boundaries, producing flickering edges and discontinuous transparency.
    We address this with a mask-guided diffusion-matting framework built on two ideas. The RGB and mask conditioning are represented as separate token streams that interact through full spatio-temporal self-attention, letting the model adaptively combine fine RGB texture with coarse mask layout rather than committing to a fixed channel-wise fusion. On top of this, we propose a novel pair of learnable embeddings, the \emph{preserve token} and the \emph{refine token}, that act as a per-frame conditioning interface and enable a sliding-window inference scheme in which already-generated mattes are propagated across chunk boundaries for local temporal consistency, complemented by a globally shared \emph{reference anchor} at every chunk to prevent error accumulation in long sequences without any additional training.
    Across standard video matting benchmarks and in-the-wild long footage, our framework outperforms prior diffusion-based methods in both fine-detail accuracy and temporal consistency, and uniquely maintains consistency across chunk boundaries on long videos.
\end{abstract}

\input{Section/1_Introduction}

\input{Section/2_RelatedWorks}
\input{Section/3_Method_ctf}
\input{Section/4_Experiments}
\input{Section/5_Conclusion}

\newpage

\section*{Acknowledgments}

{
    \small
    \bibliographystyle{plain} 
    \bibliography{main}
}

\newpage


\input{Section/X_Appendix}






\end{document}

%% file: Figures/teaser.tex
\begin{figure}[htbp]
    \centering
    \includegraphics[width=0.90\linewidth]{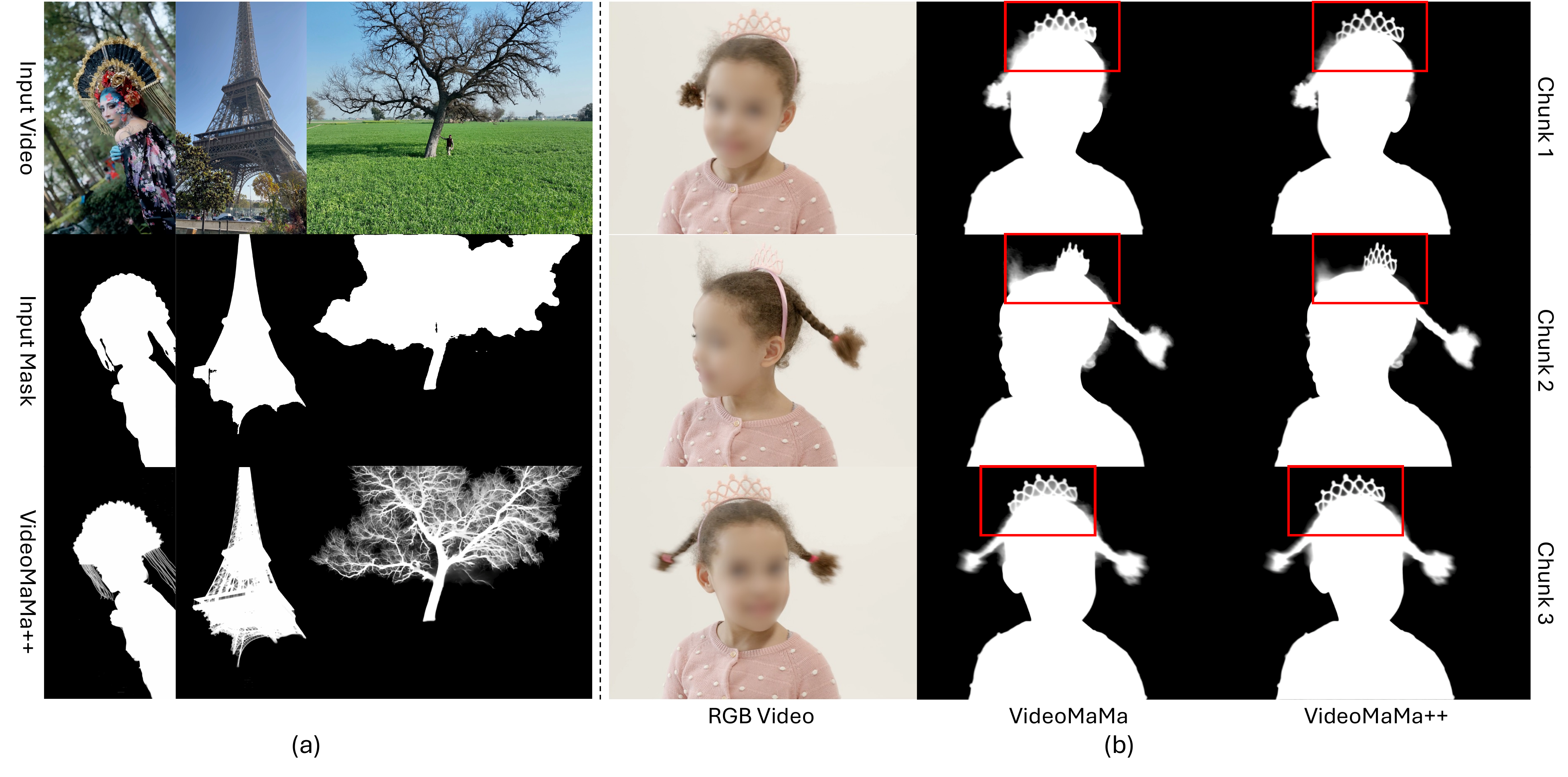}
    \vspace{-10pt}
    \caption{
    We introduce \textbf{VideoMaMa++}, a mask-guided video matting framework that recovers fine-grained details (e.g., hair strands) and scales reliably to long videos.
(a) Per-frame qualitative results on diverse in-the-wild videos.
(b) Long-video matting results: VideoMaMa~\cite{videomama}, which processes videos in chunks, fails to maintain consistency across chunk boundaries (red boxes), whereas VideoMaMa++ produces mattes that are both finely detailed and temporally consistent.
    }
    \label{fig:teaser}
    \vspace{-5pt}
\end{figure}

%% file: Section/1_Introduction.tex
\section{Introduction}
Video matting, the task of extracting foreground objects with pixel-level precision from video, is a fundamental component of video editing applications such as background replacement~\cite{zeng2025lumen, gao2025anyportal, ren2024relightful}, visual composition~\cite{dalva2024layerfusion, yang2025generative}, and relighting~\cite{chaturvedi2025synthlight, ren2024relightful, wang2025comprehensive, choi2025scribblelight}. Producing convincing composites in these settings requires alpha mattes that capture fine structures such as hair strands and motion-blurred edges, and that remain temporally consistent across the entire video.

Recent video matting methods leverage priors from video object segmentation models~\cite{maggie, matanyone}.
However, these methods still rely on dedicated matting annotations and offer limited zero-shot generalization, with most existing methods restricted to human portraits due to the scarcity of matting labels for other categories.
A more recent line of work instead leverages video diffusion priors~\cite{videomama, gvm}, transferring generative knowledge obtained from internet-scale pretraining to close the train--test gap that arises when models are trained only on synthetic matting data.
In particular, VideoMaMa~\cite{videomama} demonstrates that a mask-to-matte diffusion model trained on synthetic data alone can generalize to non-human categories and to in-the-wild video footage.

However, applying diffusion-based matting to real videos exposes a fundamental limitation that hinders general video processing. Diffusion backbones are computationally heavy and cannot process arbitrarily long sequences in a single forward pass; existing methods are therefore chunk-based, handling only 6--24 frames at a time. As a result, even ordinary videos must be split into multiple chunks and processed sequentially. When each chunk is matted independently from its coarse mask guidance, the resulting mattes drift across chunk boundaries, producing flickering edges and discontinuous transparency, as illustrated in Figure~\ref{fig:teaser} (b). The root cause is the inherent ambiguity of coarse mask guidance: a single binary mask admits many plausible fine-grained refinements at object boundaries, and independently processed chunks settle on different ones. 

In this work, we present a novel mask-guided diffusion-matting framework designed for long, high-resolution video. At its core, we represent the two guidance modalities, the RGB frames and the coarse mask, as separate token streams alongside the noisy matte tokens, and let them interact through full spatio-temporal self-attention. This design replaces the channel concatenation used in prior U-Net-based diffusion matting~\cite{videomama} and preserves each modality's distinct structure. The high-frequency RGB texture carries hair and edge cues, while the binary mask provides a coarse but spatially anchored foreground extent, and attention adaptively composes the two at every layer and spatial location. As a result, the model leans on RGB at object boundaries where the mask is unreliable, and on the mask in the interior where it confidently localizes the foreground. We instantiate this on a Diffusion Transformer (DiT)~\cite{dit, wan} with Rotary Position Embedding (RoPE)~\cite{rope}, which naturally accommodates diverse resolutions and frame counts.

To resolve the chunk-boundary inconsistency described above, we introduce a novel pair of learnable token embeddings, the \emph{preserve token} and the \emph{refine token}, a new conditioning interface for consistent video matting. The preserve token marks a frame whose guidance is already a fine matte, previously generated or ground truth, and tells the model to keep it as a reference. The refine token marks a frame with only a coarse mask and asks the model to refine it into a fine matte consistent with the preserve-tagged references via DiT's full self-attention. We employ these tokens in a sliding-window inference scheme in which each chunk overlaps a few preserve-tagged frames with the previous chunk, so that already-generated mattes propagate across chunk boundaries and maintain local temporal consistency. To further guard against error accumulation over long sequences, we additionally supply a globally reliable matte as a \emph{reference anchor} at every chunk through the same preserve token, without any additional training.

We validate our framework through extensive experiments on standard video matting benchmarks and in-the-wild footage. Our method outperforms existing video matting approaches on VideoMatte~\cite{video240k}, YoutubeMatte~\cite{matanyone}, and V-HIM60~\cite{maggie} in both fine-detail accuracy and temporal consistency, and achieves state-of-the-art performance on long-video matting where prior diffusion-based methods fail to maintain consistency across chunk boundaries. The same preserve mechanism also gives users direct control at inference time, for example to interactively correct erroneous outputs by re-tagging specific frames as refine while keeping trusted neighbors as preserve. 

Our contributions can be summarized as follows.
\begin{itemize}[noitemsep]
    \item A mask-guided diffusion-matting framework with \emph{token-level guidance representation}, where the RGB frames and the coarse mask are kept as separate token streams adaptively composed by full spatio-temporal self-attention, rather than fused via channel concatenation as in prior works.
    \item A novel \emph{preserve} and \emph{refine} token mechanism that enables chunk-consistent matting through sliding-window inference with overlap-based propagation, complemented by a globally shared \emph{reference anchor} that prevents error accumulation in long sequences without any additional training.
    \item State-of-the-art performance on standard video matting benchmarks and in-the-wild long-video footage, with substantial gains in fine-detail accuracy and temporal consistency over prior diffusion-based methods that process video chunk-wise.
\end{itemize}

%% file: Section/2_RelatedWorks.tex
\section{Related Work}

\paragraph{Video matting.}
Video matting methods can be broadly categorized into auxiliary-free and auxiliary-guided approaches~\cite{mgm-inthewild, sdmatte}. Auxiliary-free methods~\cite{lin2022robust, modnet, vmformer} typically focus on portrait matting, restricting their applicability to human-centric scenarios. Auxiliary-guided approaches relax this restriction by accepting additional conditioning at the cost of an extra annotation. Trimap-guided methods~\cite{huang2023end, otvm} require manual trimap annotations and cannot easily leverage modern segmentation models such as the SAM family~\cite{sam, sam2, sam3}. More recent mask-guided methods reduce this burden by accepting only a coarse binary mask. MaGGIe~\cite{maggie} decouples target tracking from matting but struggles to generalize to in-the-wild videos, while MatAnyone~\cite{matanyone, matanyone2} propagates target-assigned mattes through a memory-augmented architecture trained on dedicated matting datasets such as VM800~\cite{matanyone} and VMReal~\cite{matanyone2}. Despite these advances, all of these methods remain constrained by predominantly synthetic and human-centric training data, leaving in-the-wild generalization to non-human categories largely unsolved.

\paragraph{Diffusion priors for perception.}
Diffusion priors from internet-scale pretraining have been transferred to dense perception tasks. Marigold~\cite{marigold} pioneered this direction by adapting Stable Diffusion~\cite{stablediffusion} for depth estimation, and the same recipe extends to a range of video perception tasks~\cite{normalcrafter, chronoDepth, depthcrafter, geometrycrafter}, with more recent works~\cite{xu2025diffusion, zhang2026dvd} extending it to DiT-based backbones.

\vspace{-2mm}
\paragraph{Diffusion long-video inference.}
Diffusion backbones, however, are computationally heavy and limited to short chunks, so applying them to long videos requires a chunk-wise inference scheme. ChronoDepth~\cite{chronoDepth} and DepthCrafter~\cite{depthcrafter} adopt sliding-window inference with overlapping frames between adjacent chunks for local temporal smoothness, but offer no global anchor against drift. A parallel line of work tackles this through auto-regressive generation~\cite{self_forcing, diffusion_forcing, causvid, rolling_forcing, self_forcing++, longlive, flow_cache}, where each chunk is conditioned on the previous chunk via causal attention with KV-cache reuse. While effective for free-form video generation, this hidden-state propagation does not naturally compose with explicit external conditioning such as a guide mask. Our work instead propagates fully-formed reference mattes between chunks through the preserve token, keeping each chunk fully conditioned on its own user-supplied mask.

\vspace{-2mm}
\paragraph{Diffusion-based video matting.}
In the matting domain, recent diffusion-based methods transfer video diffusion priors to video matting using only synthetic data. The auxiliary-free GVM~\cite{gvm} and the mask-guided VideoMaMa~\cite{videomama} both demonstrate that this prior generalizes to non-human categories and in-the-wild footage, but both inherit the chunk-wise inference limitation of prior diffusion-based perception models. VideoMaMa, the closest mask-guided baseline, additionally fuses RGB, mask, and noise via channel concatenation, entangling the guidance modalities and limiting fine-detail recovery. Our work departs from this design by representing the two guidance modalities as separate token streams, and by introducing the preserve and refine tokens that maintain chunk-wise consistency through globally shared reference anchors and overlap-based local smoothing.

%% file: Section/3_Method_ctf.tex
\section{Method}
\input{Figures/main_architecture}

\subsection{Problem Formulation and Overview}
Video matting is mathematically defined by the alpha compositing equation,
\begin{equation}
    I_t = \alpha_t F_t + (1 - \alpha_t) B_t, \quad t = 1, \dots, T,
\end{equation}
where $t$ denotes the frame index, $T$ is the total number of frames in the video, $I_t$ is the observed image at frame $t$, $F_t$ is the foreground, $B_t$ is the background, and $\alpha_t \in [0,1]$ is the alpha matte representing pixel-level opacity. Unlike binary segmentation masks $M \in \{0, 1\}$, alpha mattes capture fine-grained details such as hair strands, motion blur, and semi-transparent regions, which are critical for realistic video compositing.

We adopt a mask-guided formulation that decomposes video matting into two subproblems: foreground tracking and boundary refinement. Tracking a target object across a video is itself a demanding task, and entangling it with matte generation forces a single model to simultaneously localize objects and resolve pixel-level matting. By conditioning on a binary mask, we decouple these two challenges and allow our model to focus on generating fine-grained matting details such as hair strands and motion blur, while inheriting localization capabilities from off-the-shelf segmentation models~\cite{sam2, sam3, xmem, Cutie}.

\label{sec:pipeline}
We design an end-to-end pipeline for long-video matting that takes a video and a single user prompt on the first frame as input, and produces a temporally consistent alpha matte across the entire sequence. The pipeline operates in three stages, illustrated in Figure~\ref{fig:main_architecture}. First, we obtain a per-frame foreground signal by propagating the user prompt across all frames using an off-the-shelf video segmentation model such as Segment Anything~\cite{sam2, sam3}. Second, the resulting binary mask track is passed to our Diffusion Transformer (DiT) backbone together with the RGB frames to generate fine-grained alpha mattes (Section~\ref{sec:dit_prior}). Finally, long sequences are handled by a chunk-wise inference scheme equipped with preserve and refine tokens (Section~\ref{sec:preserve_refine}), which propagates information across chunk boundaries to maintain temporal consistency throughout the video.

\subsection{Repurposing Diffusion Transformer for Video Matting}
\label{sec:dit_prior}
Video matting annotations are scarce and limited in quantity and category coverage. Generalizing to real-world footage therefore requires a strong prior over natural video content. We build our framework on a video DiT~\cite{dit, wan} backbone, which provides a rich generative prior learned from internet-scale video pretraining and offers several architectural properties well-suited to video matting.

The input video $V$, the guide mask track $M$, and, during training, the ground-truth alpha matte $\alpha$ are each encoded into the latent space by a shared, frozen VAE encoder $\mathcal{E}$, producing $z_V$, $z_M$, and $z_\alpha$ respectively. We sample Gaussian noise $\epsilon \sim \mathcal{N}(0, I)$ with the same shape as $z_\alpha$, and concatenate $z_V$, $z_M$, and $\epsilon$ along the token dimension as input to the DiT backbone:
\begin{equation}
    \hat{z}_{\alpha} = \mathcal{F}_\text{DiT}\big(\text{concat}(z_V,\, z_M,\, \epsilon)\big),
    \label{eq:single_step_model}
\end{equation}
where $\mathcal{F}_\text{DiT}(\cdot)$ denotes our adapted DiT and $\text{concat}(\cdot)$ denotes concatenation. Unlike UNet-based matting models~\cite{videomama, svd} that fuse the three modalities along the channel dimension, our token-level concatenation lets each modality occupy its own token stream and preserves modality-specific information throughout the network. The final alpha matte video is obtained by decoding the predicted latents through the VAE decoder, $\hat{\alpha} = \mathcal{D}(\hat{z}_\alpha)$. We initialize from the pretrained DiT~\cite{wan} and, instead of its original flow-matching objective, supervise the decoded predictions directly in pixel space (see Appendix~\ref{sec:training_details}). Although training follows the standard multi-step schedule, we find a single denoising step at inference sufficient to produce high-quality mattes, and adopt it as our default for efficiency.

\subsection{Preserve and Refine Tokens}
\label{sec:preserve_refine}
The formulation above operates on a single clip, while video matting in practice requires handling long sequences. DiT backbones are computationally heavy and cannot process long videos in a single forward pass. A natural solution is chunk-wise inference, where the video is split into fixed-length clips and each clip is processed independently. However, when each chunk is refined from its guidance in isolation, the resulting mattes drift across chunk boundaries, producing flickering edges and discontinuous transparency.

To address this, we introduce two learnable token embeddings, a \emph{preserve token} and a \emph{refine token}, which assign each frame a per-frame mode that determines whether its matte is carried through unchanged or generated from coarse guidance. The preserve token marks anchor frames whose alpha mattes should be retained, while the refine token marks frames that should be converted from coarse guidance into fine-grained alpha mattes. DiT's full self-attention propagates information from preserve frames to refine frames, yielding mattes that remain consistent across chunks.

\paragrapht{Why not auto-regressive diffusion?}
We chose to propagate fully-formed reference mattes through the preserve token rather than reuse the previous chunk's key-value (KV) cache as in auto-regressive diffusion approaches~\cite{self_forcing, causvid}. Hidden-state propagation implicitly turns the model into a propagation-based matting framework, where each chunk's prediction is determined more by the previous chunk's internal state than by its own mask guidance. This re-entangles localization and fine-detail recovery, the two subproblems we explicitly decoupled by adopting mask-guided conditioning. The preserve token instead exposes the previous chunk's contribution as an explicit alpha matte, so each chunk remains fully grounded in its own mask guidance while still receiving long-range temporal context through self-attention.

\paragrapht{Training.}
We train both tokens jointly within a single clip. For each training clip, we randomly select a subset of frames as preserve frames, replace their guide input with the ground-truth alpha matte, and add the preserve token embedding to those frames. The remaining frames receive the refine token together with the standard binary mask guide input. The model is supervised against the ground-truth matte at all frames through the pixel-level supervision, learning to reconstruct the supplied matte at preserve frames and to generate fine-grained mattes from coarse guidance at refine frames. Joint training within a single clip is essential, as it forces the DiT to learn a unified representation in which preserve and refine frames interact through self-attention, rather than treating them as two disjoint tasks.

\subsection{Temporally Consistent Inference}
\label{sec:temporal_consistent_inference}
We deploy the preserve and refine tokens at inference through two complementary mechanisms: a sliding-window scheme that propagates mattes between adjacent chunks for local consistency, and a reference anchor that supplies a globally reliable matte at every chunk to prevent long-range drift.

\paragrapht{Sliding window.}
At inference, we apply a sliding-window scheme in which consecutive chunks share an overlap region. The overlapping frames receive the previous chunk's predicted matte as input together with the preserve token, so that their mattes are passed through unchanged, while the remaining frames receive the guide signal together with the refine token. Although the preserve token is applied only to the overlap frames, DiT's full attention allows them to condition the generation of every other frame in the chunk, transferring temporal context across the chunk boundary and producing chunk-consistent mattes.

\paragrapht{Reference anchor.}
The sliding-window scheme alone is vulnerable to error accumulation, since a corrupted overlap frame, e.g., due to occlusion, tracking drift, or other video artifacts, propagates through the preserve token to all subsequent chunks and cascades local failures through the rest of the video. To prevent this, we introduce a \emph{reference anchor}, a globally reliable frame whose matte is supplied as preserve input alongside the local overlap frames at every chunk, providing a stable, error-free signal that the model can attend to throughout inference while the overlap frames continue to carry short-range temporal context. Crucially, this capability emerges directly from the preserve token without any additional training: because the preserve token already learns to treat its inputs as anchors, any user-supplied matte can serve as an anchor for the rest of the video.

This design gives users explicit control over which frame anchors the inference. By default, we reuse the matte predicted at the first chunk as the reference anchor throughout the video, which we find sufficient for most sequences. When the first frame is unreliable due to heavy occlusion, motion blur, or partial visibility, the user can instead select a later frame with cleaner foreground appearance as the reference, allowing the entire sequence to be conditioned on the most informative anchor available.

%% file: Figures/main_architecture.tex
\begin{figure}[htbp]
    \centering
    \includegraphics[width=\linewidth]{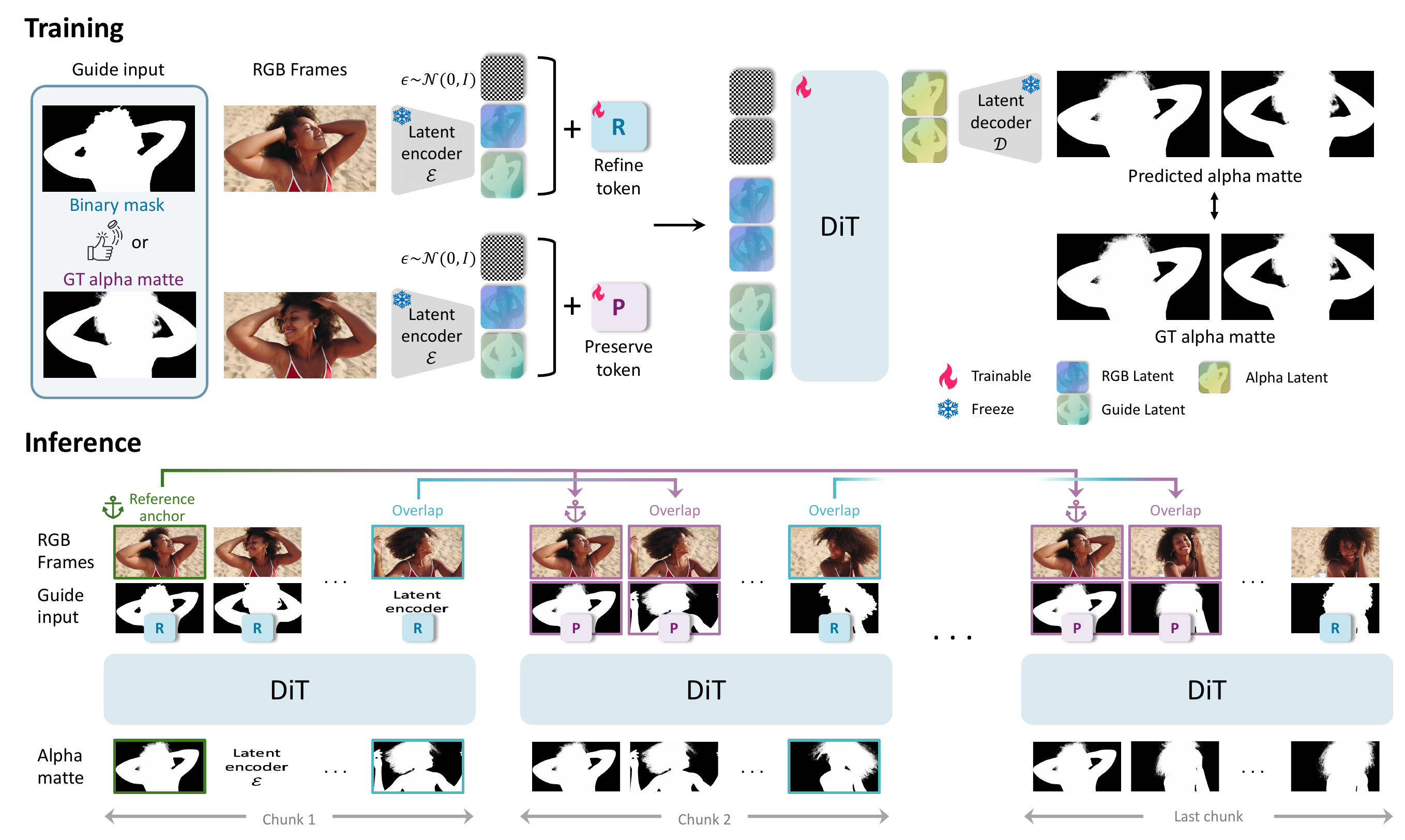}
    \caption{\textbf{Overview of the VideoMaMa++ training and inference.} \textbf{(Top)} During training, RGB frames are paired with either a binary mask or the ground-truth alpha matte as guide input, encoded into separate token streams by a frozen VAE encoder $\mathcal{E}$, and tagged with a refine token (\textbf{R}) or preserve token (\textbf{P}). The DiT processes both modes jointly within a single clip, supervised against the ground-truth matte. \textbf{(Bottom)} At inference, long videos are processed in a sliding-window manner. Each chunk receives the refine token on its non-overlapping frames and the preserve token on its overlap frames, which carry the previous chunk's predicted matte. A globally selected reference anchor frame is additionally supplied as preserve input to every chunk, providing a stable anchor that prevents error accumulation across the sequence.}
    \label{fig:main_architecture}
    \vspace{-10pt}
\end{figure}

%% file: Section/4_Experiments.tex
%
%

\section{Experiments}

\subsection{Implementation Details}
We build VideoMaMa++ on top of the Wan 2.1 (1.3B)~\cite{wan} model, training on a combination of image and video matting datasets. To preserve the generative prior, we adapt the DiT to the video matting task using LoRA~\cite{lora} rather than fine-tuning the full backbone. Training uses 3-frame clips at $1024 \times 1024$ resolution, which provides sufficient capacity for both spatial detail and temporal modeling. At inference, we process the video in chunks of 6 frames with a 2-frame overlap between consecutive chunks, where one of the overlap frames is the reference anchor.
Although the training and inference settings differ in both frame count and chunking behavior, the model generalizes across them due to RoPE's~\cite{rope} relative position encoding, which removes the dependence on absolute resolution and clip length present in fixed positional embeddings.

\subsection{Quantitative Evaluations}
\label{sec:Quantitative}

\input{Tables/Single_chunk_simple}
We evaluate VideoMaMa++ on three benchmarks: VideoMatte240K~\cite{video240k}, YouTubeMatte~\cite{matanyone}, and V-HIM60 (Hard)~\cite{maggie}.
All mask-guided methods, including ours, receive identical guidance masks produced by a single procedure (Algorithm~\ref{alg:mask_protocol}). The binarized first-frame ground-truth matte prompts SAM3~\cite{sam3} on a duplicated first frame (frames $\{0,0,1,2,\dots\}$), and SAM3 propagates it across the video, so every guidance mask, including that of the first frame, is a SAM3 prediction rather than the binarized ground truth, and no method receives a privileged mask.
For one YouTubeMatte sequence, a sheet of paper that belongs to the foreground enters mid-video, so all methods share a track initialized at a later frame in which the paper is visible; VideoMaMa++ remains the best method under first-frame initialization as well (Appendix~\ref{sec:youtubematte_remask}).

As reported in Table~\ref{tab:quan}, VideoMaMa++ achieves the best performance across all metrics on every benchmark. The improvements are particularly pronounced on Gradient~\cite{rhemann2009perceptually} and dtSSD~\cite{dtssd}, which respectively measure spatial boundary sharpness and temporal smoothness. The Gradient gains confirm that the DiT-based generative prior recovers fine-grained boundary detail more accurately than prior mask-guided methods, while the dtSSD improvements indicate that the preserve and refine token mechanism produces temporally consistent mattes despite chunk-wise inference. 
We provide a further analysis of discontinuities at chunk boundaries in Appendix~\ref{sec:app_seam}.

\vspace{-3mm}
\subsection{Qualitative Evaluations}
\label{sec:qualitative}
\vspace{-2mm}
\input{Figures/main_qual_long}
We compare VideoMaMa++ against representative video matting baselines, including MGM in-the-wild~\cite{mgm-inthewild}, MaGGIe~\cite{maggie}, MatAnyone~2~\cite{matanyone2}, and VideoMaMa~\cite{videomama}, on in-the-wild videos.
Figure~\ref{fig:main_qual_long} shows long videos that diffusion models (VideoMaMa and ours) cannot process in a single chunk. As shown in the figure, the comparison methods either fail to capture fine-grained details or produce flickering results across chunks that are inconsistent over the full video.
In contrast, VideoMaMa++ produces mattes that retain fine-grained boundary detail consistently throughout the sequence and remain stable across distant frames, demonstrating that our DiT-based generative prior, combined with the preserve and refine token mechanism, generalizes effectively to long-form footage well beyond the length seen during training.
We refer the reader to the supplementary material for additional comparisons that more clearly demonstrate temporal consistency in motion.

We further quantify this observation with an annotation-free test for chunk-boundary discontinuities (Appendix~\ref{sec:app_seam}), which detects spikes in the frame-to-frame matte change that recur at chunk seams. With both models run at identical seams, VideoMaMa's matte change spikes at seams in-the-wild clips, whereas VideoMaMa++ stays at chance level. On YouTubeMatte, where ground truth is available, the same test detects VideoMaMa's discontinuity and finds none in the ground-truth mattes, supporting its validity. Per clip, VideoMaMa's seam contrast exceeds that of VideoMaMa++ at identical boundaries, and these conclusions are stable across difference statistics and detrending windows. Since the test targets only chunk-boundary discontinuities, it applies only to chunk-wise methods, which is why we compare against VideoMaMa.

\paragrapht{Interactive refinement.} 
\input{Figures/Supplementary/interactive_qual}
The preserve and refine tokens also enable selective refinement of individual frames within a chunk. Given an initial prediction, we identify the frame with the largest error against the ground-truth matte, mark it with the refine token, and mark the remaining frames with the preserve token. Re-running the model under this configuration regenerates the marked frame while attending to the others as anchors, producing a refined matte that draws on temporal context from the rest of the chunk. As illustrated in Figure~\ref{fig:interactive_qual}, this interactive procedure consistently improves matting quality, even though no new guide signal is provided. Quantitative results for interactive refinement are reported in the appendix.

\vspace{-2mm}
\subsection{Ablation Studies}
\vspace{2mm}

\paragrapht{Reference anchor.}
\input{Tables/abl_anchor}
We ablate the reference anchor by comparing two sliding-window inference settings: \emph{w/o r. anchor}, which uses previous chunk's matte on the local overlap frames with preserve tokens, and \emph{w/ r. anchor}, our full setting that additionally supplies the first-chunk matte as a reference anchor providing a global preserve input. As reported in Table~\ref{tab:abl_first_frame_anchor}, adding the reference anchor yields consistent improvements across all six metrics on both benchmarks, with the largest gains on boundary-sensitive measures such as Gradient. This confirms that propagating preserve information through local overlaps alone leaves the model vulnerable to error accumulation, and that the global anchor effectively stabilizes long-range inference while preserving the cross-chunk information transfer enabled by the preserve token.

The aggregate metrics in Table~\ref{tab:abl_first_frame_anchor} measure average matting quality but do not reveal how that quality evolves as the sliding window advances along the sequence. To probe error accumulation directly, we re-aggregate MAD by chunk index, and report the relative change with respect to the first chunk.
As shown in Figure~\ref{fig:abl_error_propagation}, the \emph{w/o r. anchor} variant exhibits a steadily rising curve on both benchmarks, with MAD growing markedly by the final chunk. In contrast, the \emph{w/ r. anchor} variant remains substantially flatter throughout the sequence, confirming that the reference anchor effectively suppresses error propagation across long videos.

\paragrapht{Choice and quality of the reference anchor.}
\begin{table}[t]
    \centering
    \caption{\textbf{Reference anchor analysis.}
    We evaluate (a) the anchor frame within the first chunk and re-anchoring every 3 chunks (Refresh), and (b) a corrupted anchor frame: dilated (D) or eroded (E) guidance mask, motion blur (B), and full occlusion (Occ.), with kernel sizes in px. In (b), the corrupted first frame is excluded from evaluation. All evaluations are performed on the \textbf{V-Him60 (Hard)} benchmark. The \textbf{best} results in (a) are highlighted.
    }
    \label{tab:anchor}
    \begin{adjustbox}{width=\linewidth}
        \begin{tabular}{@{}l cccc c cccc ccc c@{}}
            \toprule
             & \multicolumn{4}{c}{\textbf{(a) Anchor choice}} & \multicolumn{9}{c}{\textbf{(b) Corrupted anchor frame}} \\
            \cmidrule(lr){2-5} \cmidrule(l){6-14}
             & & & & & & \multicolumn{4}{c}{Poor mask} & \multicolumn{3}{c}{Motion blur} & \\
            \cmidrule(lr){7-10} \cmidrule(lr){11-13}
            Metrics & First & Random & Last & Refresh & Clean & D5 & D21 & E5 & E21 & B7 & B15 & B25 & Occ. \\
            \midrule
            MAD $\downarrow$   & \textbf{1.69} & 1.70 & 1.72 & 1.70 & 1.71 & 1.71 & 1.73 & 1.73 & 1.74 & 1.74 & 1.74 & 1.75 & 1.76 \\
            MSE $\downarrow$   & \textbf{0.45} & \textbf{0.45} & \textbf{0.45} & 0.46 & 0.46 & 0.46 & 0.46 & 0.46 & 0.48 & 0.48 & 0.48 & 0.48 & 0.48 \\
            Grad $\downarrow$  & \textbf{2.92} & 2.94 & 2.95 & 2.94 & 2.95 & 2.94 & 2.94 & 2.97 & 2.95 & 3.04 & 3.05 & 3.05 & 3.06 \\
            dtSSD $\downarrow$ & \textbf{1.98} & 1.99 & 2.00 & 1.99 & 1.99 & 1.99 & 1.98 & 2.00 & 1.99 & 2.02 & 2.02 & 2.03 & 2.03 \\
            Conn $\downarrow$  & \textbf{2.72} & 2.74 & 2.76 & --   & 2.75 & 2.75 & 2.78 & 2.78 & 2.81 & 2.81 & 2.81 & 2.83 & 2.84 \\
            \bottomrule
        \end{tabular}
    \end{adjustbox}
    \vspace{-15pt}
\end{table}

The first frame may not represent later frames well, for example when the subject is partially occluded or only enters the frame later. Using a random frame or the last frame of the first chunk as the anchor instead yields nearly unchanged performance across all metrics (Table~\ref{tab:anchor}a), indicating that any frame of the first chunk serves as an effective anchor. Re-anchoring to a recent frame every few chunks likewise matches the default setting, so a first-frame anchor does not become stale on V-HIM60 despite its substantial pose, scale, and appearance variation; refreshing nonetheless remains available for footage with more drastic changes.
 
We further corrupt the input of the anchor frame (Table~\ref{tab:anchor}b) with a poor guidance mask (erosion or dilation), horizontal motion blur on the RGB frame, or a real occluder composited over nearly the entire subject, whose region is also removed from the guidance mask (Appendix~\ref{sec:app_anchor}). Because the anchor is the model's own first-frame prediction, the degraded anchor matte propagates to all later chunks. Nevertheless, performance degrades only marginally under every corruption. Full occlusion is the most severe case, followed by motion blur and poor guidance masks, and even a fully occluded anchor leaves the model close to its clean-anchor performance.

\paragrapht{Effect of preserve token at chunk boundaries.}
\input{Tables/abl_overlap_diff}
We design a targeted experiment to verify that the preserve and refine tokens produce consistent mattes across chunk boundaries. Using a chunk size of 6 with an overlap of 2 frames, each overlap frame is generated twice: once as part of the preceding chunk and once as part of the following chunk. We measure the discrepancy between these two predictions for the same frame, which directly reflects how much the matte drifts when the model crosses a chunk boundary. The two settings differ only in how the overlap frames are treated by the second chunk: \emph{w/o preserve} regenerates them independently from the coarse mask guidance, whereas \emph{w/ preserve} marks them with the preserve token and conditions on the matte produced by the preceding chunk. To isolate the effect of the preserve token, the reference anchor is disabled in both settings. As reported in Table~\ref{tab:overlap_diff}, \emph{w/ preserve} reduces both MAD and Gradient discrepancy by a large margin, confirming that the preserve token effectively transfers cross-chunk information and yields mattes that align across chunk boundaries rather than drifting into independently chosen refinements.

\paragrapht{Robustness to mask generators.}
\input{Tables/abl_mask_diverse}
To evaluate the robustness of our model to diverse binary mask sources, we measure performance on V-HIM60~\cite{maggie} using XMem~\cite{xmem}, Cutie~\cite{Cutie}, SAM2~\cite{sam2}, and SAM3~\cite{sam3} as binary mask generators in Table~\ref{tab:abl_mask_generators}. The results show that the model maintains consistent performance across all mask generators, demonstrating its robustness to the choice of binary mask input.
Beyond the choice of generator, the model also tolerates local mask errors: when the guidance mask is wrong on individual frames of a chunk, such as missing thin structures or drift under motion blur, cross-frame attention to neighboring frames recovers the matte. If the tracker instead locks onto a wrong object for an entire sequence, the matte follows that track, a limitation shared by all mask-guided methods.

\paragrapht{Continuous guidance via SAM logits.}
\input{Figures/sam_logit_binary_comparison}
We investigate replacing the binary guide mask with SAM's raw logit map, which preserves the soft confidence information discarded by binarization (Figure~\ref{fig:sam_logit_binary_comparison}). The structures most affected by this collapse, such as hair strands, thin wires, and motion-blurred edges, exhibit below-threshold logits that are removed entirely once the mask is binarized. We replace the guide mask track $M$ in Section~\ref{sec:dit_prior} with a SAM logit map, encoded by the same VAE into a latent that substitutes for $z_M$ in Eq.~\ref{eq:single_step_model}, and retrain the model end-to-end with this logit conditioning rather than swapping inputs only at inference. As reported in Table~\ref{tab:abl_samlogit}, the logit-conditioned variant outperforms its binary counterpart across all four metrics, confirming that soft confidence carries signals critical for fine-detail recovery. However, this gain does not transfer to in-the-wild videos, where the model overfits to SAM's specific logit distribution. We therefore adopt binary masks as the default guidance.
\input{Tables/ablations}

\paragrapht{Channel vs token concatenation.}
We compare two strategies for combining the RGB, mask, and noise inputs: channel concatenation, which stacks all three modalities along the channel axis into a single input tensor, and token concatenation, which appends each modality as a separate token stream. The channel-concat variant uses an expanded patch embedding to ingest the stacked input while keeping the rest of the DiT backbone identical to our model. Channel concatenation reduces the token sequence length and therefore the compute cost of self-attention, but as shown in Table~\ref{tab:abl_input}, it consistently underperforms token concatenation across all metrics. We attribute this to the loss of modality-specific information when the inputs are collapsed into a single tensor at the input stage.

\paragrapht{Number of denoising steps.}
\input{Tables/abl_denoising_step}
We evaluate how the number of denoising steps at inference affects matting quality. Following the observation of GenPercept~\cite{genpercept} in monocular depth estimation, we find that a single denoising step is sufficient for the diffusion model to convert a binary mask into a high-quality alpha matte. As reported in Table~\ref{tab:abl_denoising_steps}, additional denoising steps lead the model to over-refine the matte, degrading both boundary sharpness and overall accuracy. We therefore adopt single-step inference as the default, which also yields a substantial efficiency gain for large-scale video processing (Appendix~\ref{sec:cost}).

\paragrapht{Chunk size.}
\input{Tables/abl_chunksize}
Table~\ref{tab:abl_chunk_size} evaluates chunk sizes on V-HIM60~\cite{maggie}, with the overlap fixed at 2 frames across all settings. A 6-frame chunk outperforms both smaller (3) and larger (12) settings. Small chunks restrict the temporal context available for DiT's self-attention, whereas large chunks dilute the preserve frames' influence, reducing stability. A 6-frame chunk optimally balances temporal context with an effective preserve signal, serving as our default.

\paragrapht{LoRA rank.}
\input{Tables/abl_lora}
We ablate the LoRA rank used to adapt the pretrained DiT for video matting. As reported in Table~\ref{tab:abl_lora_rank}, performance improves monotonically as the rank increases, with rank 512 achieving the best results across all six metrics. The lower-rank variants (64 and 128) underperform on boundary-sensitive measures such as Gradient and MAD-T, indicating that mask-to-matte refinement requires sufficient adapter capacity to learn the fine-grained boundary structure absent from the input mask. We therefore adopt rank 512 as our default, which provides the best matting quality while remaining substantially lighter than full fine-tuning of the DiT backbone to preserve generative prior.


%% file: Tables/Single_chunk_simple.tex
\begin{table}[t]
    \centering
    \caption{\textbf{Quantitative comparison of video matting methods on various benchmarks.}
    }
    \label{tab:quan}
    \begin{adjustbox}{width=\linewidth}
        \begin{tabular}{ll ccccccccc}
            \toprule
            \textbf{{Dataset}} & Metrics & SAM3 & MGM & MGM-IW & MaGGIe & MaGGIe-FT & MatAnyone & MatAnyone 2 & VideoMaMa & Ours \\
            \midrule
            {\textbf{VideoMatte}}           & MAD $\downarrow$   & 5.05 & 11.28 & 5.01 & 4.26 & 4.44 & 4.17 & \underline{4.04} & 4.34 & \textbf{3.86} \\
            {\textbf{(1920 $\times$ 1080)}} & MSE $\downarrow$   & 0.94 & 3.23 & 0.35 & 0.27 & 0.33 & 0.27 & \underline{0.24} & 0.27 & \textbf{0.12} \\
                                                   & Grad $\downarrow$  & 12.02 & 6.33 & 5.01 & 3.61 & 4.53 & 3.59 & \underline{3.13} & 4.06 & \textbf{1.79} \\
                                                   & dtSSD $\downarrow$ & 3.11 & 2.11 & 1.38 & 1.23 & 1.35 & 1.17 & \underline{1.13} & 1.21 & \textbf{0.82} \\
            \midrule
            {\textbf{YoutubeMatte}}         & MAD $\downarrow$   & 2.35 & 3.59 & 2.41 & 1.67 & 1.52 & 1.74 & 1.52 & \underline{1.41} & \textbf{1.15} \\
            {\textbf{(1920 $\times$ 1080)}} & MSE $\downarrow$   & 1.11 & 0.53 & 0.38 & 0.50 & 0.37 & 0.53 & 0.44 & \underline{0.30} & \textbf{0.23} \\
                                                   & Grad $\downarrow$  & 14.63 & 4.27 & 4.25 & 5.11 & 5.36 & 7.37 & 6.46 & \underline{4.08} & \textbf{3.27} \\
                                                   & dtSSD $\downarrow$ & 3.28 & 1.57 & 1.37 & 1.46 & 1.47 & 1.56 & 1.48 & \underline{1.31} & \textbf{1.17} \\
            \midrule
            {\textbf{V-HIM60}}              & MAD $\downarrow$   & 3.91 & 66.31 & 2.89 & 2.72 & 2.81 & 4.04 & 4.73 & \underline{2.30} & \textbf{1.69} \\
            {\textbf{(Hard)}}               & MSE $\downarrow$   & 2.09 & 49.02 & 0.84 & 1.03 & 1.03 & 2.29 & 2.93 & \underline{0.75} & \textbf{0.45} \\
                                                   & Grad $\downarrow$  & 18.37 & 17.53 & \underline{4.86} & 6.05 & 6.53 & 7.04 & 6.69 & 5.04 & \textbf{2.92} \\
                                                   & dtSSD $\downarrow$ & 5.08 & 12.73 & 2.74 & 3.18 & 3.12 & 4.08 & 3.70 & \underline{2.51} & \textbf{1.98} \\
                                                   & Conn $\downarrow$  & 6.97 & 114.70 & 4.08 & 4.72 & 4.84 & 7.10 & 8.06 & \underline{3.83} & \textbf{2.72} \\
            \bottomrule
        \end{tabular}
    \end{adjustbox}
    \vspace{-15pt}
\end{table}

%% file: Figures/main_qual_long.tex
\begin{figure}[t]
    \centering
    \includegraphics[width=\linewidth]{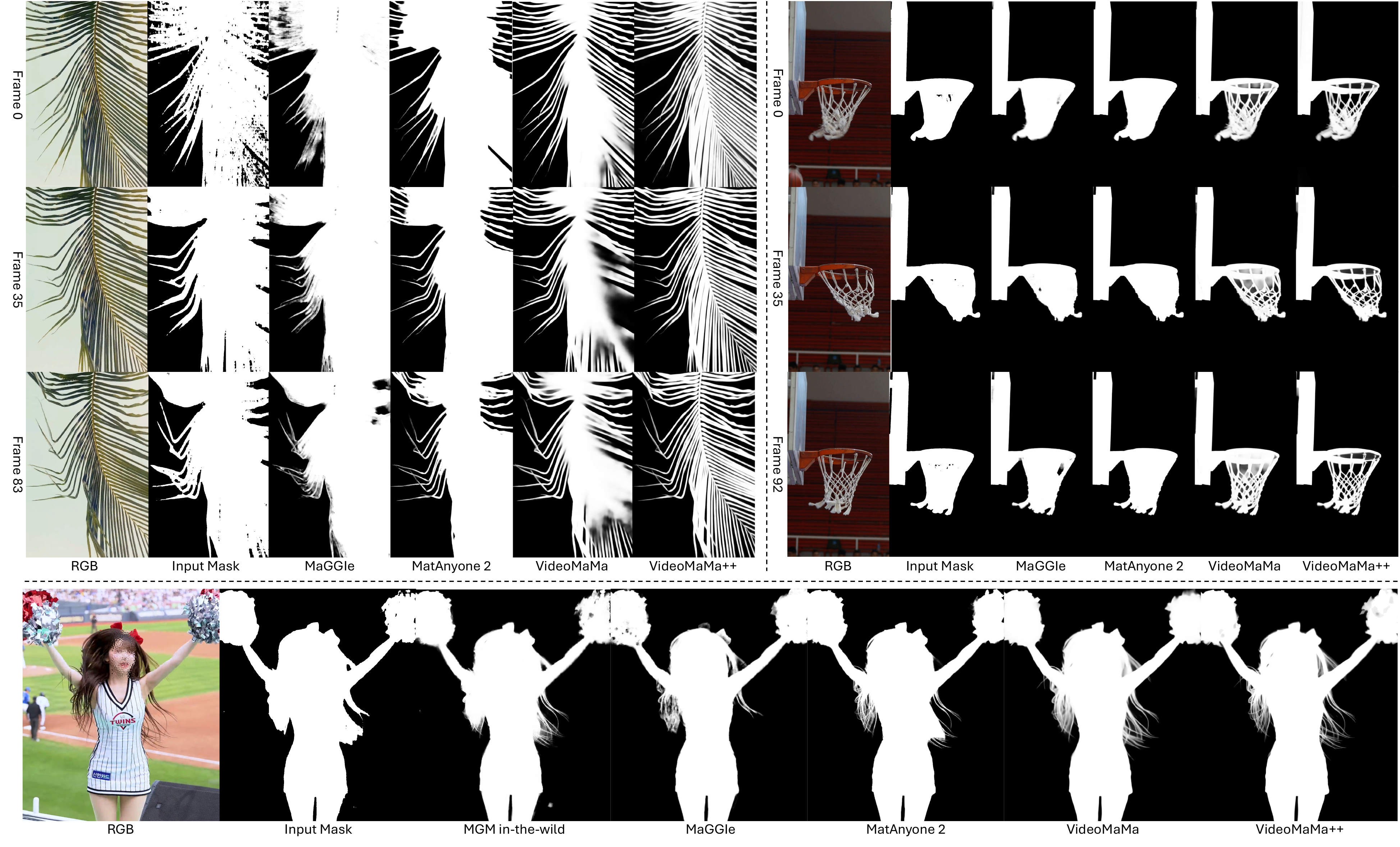}
    \vspace{-15pt}
    \caption{\textbf{Long-video temporal consistency.} Qualitative comparison between VideoMaMa++ and other methods. The comparison methods exhibits visible flickering at chunk boundaries or struggles to capture fine-grained details, while VideoMaMa++ produces smooth transitions across chunks while still capturing fine-grained details that emerge in later chunks.}
    \label{fig:main_qual_long}
    \vspace{-10pt}
\end{figure}

%% file: Figures/Supplementary/interactive_qual.tex

\begin{figure}[t]
    \centering
    \begin{subfigure}{0.48\linewidth}
        \centering
        \includegraphics[width=\linewidth]{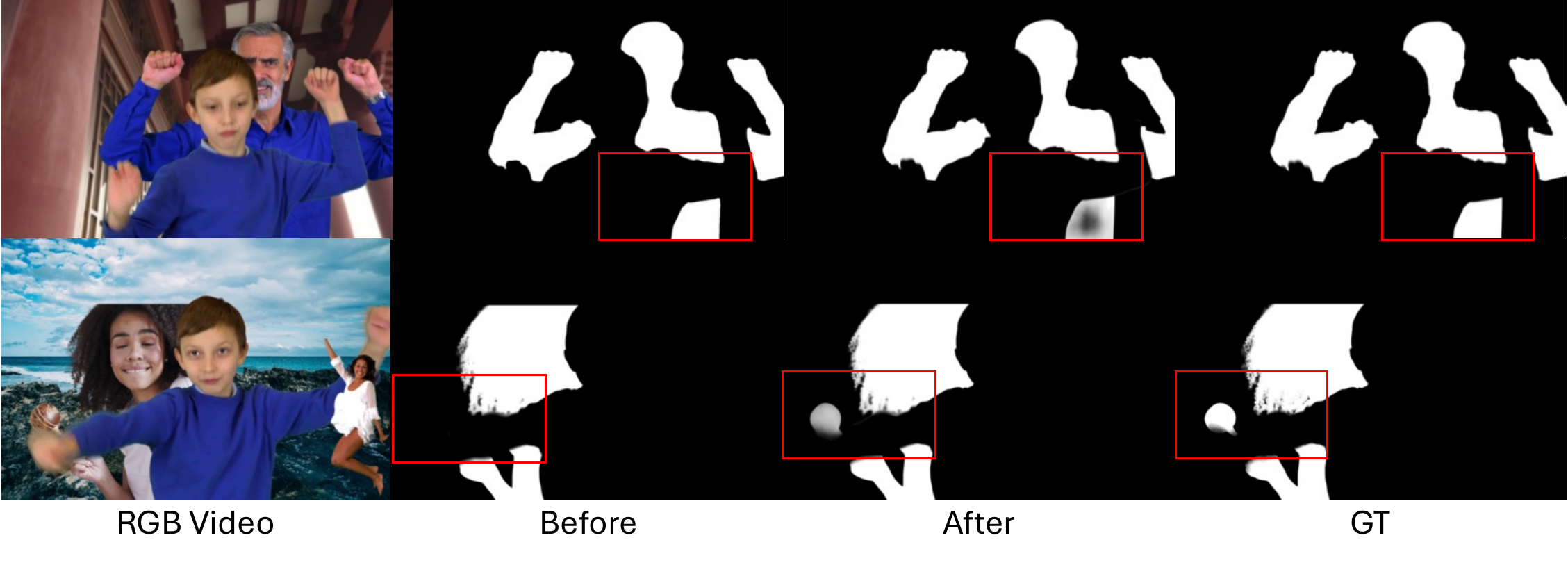}
        \caption{\textbf{Interactive refinement examples.}
        }
        \label{fig:interactive_qual}
    \end{subfigure}
    \begin{subfigure}{0.5\linewidth}
        \centering
        \includegraphics[width=\linewidth]{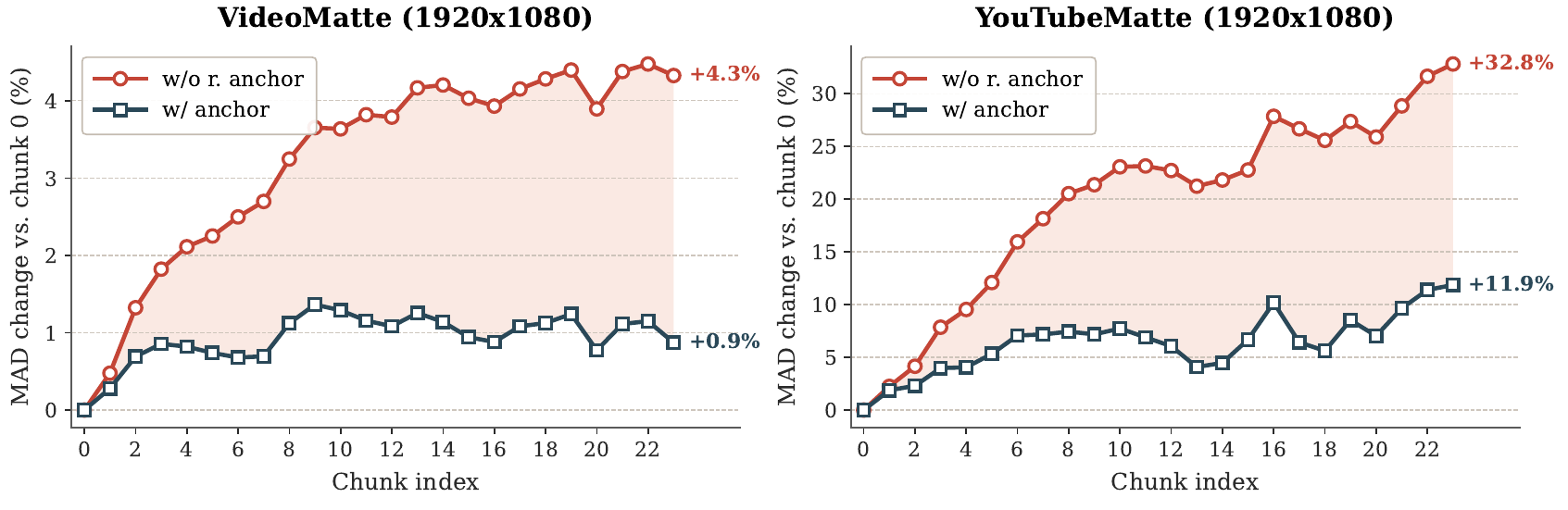}
        \caption{\textbf{MAD across chunk index (relative to chunk 0).}}
        \label{fig:abl_error_propagation}
    \end{subfigure}
    \hfill 
    \caption{\textbf{(Left)} Interactive refinement examples. \textbf{(Right)} MAD across chunk index.}
    \label{fig:combined_figure}
    \vspace{-15pt}
\end{figure}

%% file: Tables/abl_anchor.tex
\begin{wraptable}{r}{0.5\linewidth}
    \centering
    \vspace{-\baselineskip}
    \caption{\textbf{Ablation study on reference anchor.} 
    }
    \label{tab:abl_first_frame_anchor}
    \vspace{-5pt}
    \resizebox{\linewidth}{!}{%
    \begin{tabular}{@{}l cccccc@{}}
      \toprule
      \textbf{Setting} & \textbf{MAD} $\downarrow$ & \textbf{MAD-T} $\downarrow$ & \textbf{MSE} $\downarrow$ & \textbf{GRAD} $\downarrow$ & \textbf{CONN} $\downarrow$ & \textbf{DTSSD} $\downarrow$ \\
      \midrule
      \multicolumn{7}{@{}l}{\textbf{{VideoMatte (1920 $\times$ 1080)}}} \\
      \midrule
      w/o r. anchor  & 3.9505  & 4.7366  & 0.1518  & 2.2398  & 1.2817  & 0.8576 \\
      w/ r. anchor   & 3.8633  & 4.6354  & 0.1240  & 1.7887  & 1.0755  & 0.8206 \\
      \midrule
      \multicolumn{7}{@{}l}{\textbf{{YoutubeMatte (1920 $\times$ 1080)}}} \\
      \midrule
      w/o r. anchor  & 1.2691  & 31.6900  & 0.2861  & 4.4151  & 2.3294  & 1.1803 \\
      w/ r. anchor   & 1.1526  & 29.4655  & 0.2288  & 3.2741  & 2.0606  & 1.1663 \\
      \bottomrule
    \end{tabular}%
    }
    \vspace{-10pt}
\end{wraptable}

%% file: Tables/abl_overlap_diff.tex
\begin{wraptable}{r}{0.4\linewidth}
    \centering
    \vspace{-10pt}
    \caption{\textbf{Chunk boundary consistency.} 
    }
    \label{tab:overlap_diff}
    \vspace{-5pt}
    \resizebox{\linewidth}{!}{%
    \begingroup
    \renewcommand{\arraystretch}{0.85} 
    \setlength{\tabcolsep}{10pt} 
    \begin{tabular}{@{}l cc@{}}
      \toprule
      \textbf{Method} & \textbf{MAD} $\downarrow$ & \textbf{Gradient} $\downarrow$ \\
      \midrule
      w/o preserve  & 0.7128 & 0.6811 \\
      w/ preserve   & \textbf{0.2583} & \textbf{0.1235} \\
      \bottomrule
    \end{tabular}%
    \endgroup
    }
    \vspace{-15pt}
\end{wraptable}

%% file: Tables/abl_mask_diverse.tex
\begin{wraptable}{r}{0.43\linewidth}
    \centering
    \vspace{-10pt}
    \caption{\textbf{Robustness to mask generators.}}
    \label{tab:abl_mask_generators}
    \vspace{-5pt}
    \resizebox{\linewidth}{!}{%
    \begingroup
    \renewcommand{\arraystretch}{0.85} 
    \setlength{\tabcolsep}{8pt} 
    \begin{tabular}{@{}l cccc@{}}
      \toprule
      & \multicolumn{2}{c}{\textbf{MAD} $\downarrow$} & \multicolumn{2}{c}{\textbf{Gradient} $\downarrow$} \\
      \cmidrule(lr){2-3} \cmidrule(lr){4-5}
      \textbf{Input Mask} & \textbf{Input} & \textbf{Refined} & \textbf{Input} & \textbf{Refined} \\
      \midrule
      XMem  & 5.5587 & 3.1119 & 17.7265 & 3.6025 \\
      Cutie & 3.6998 & 1.9430 & 13.5395 & \textbf{2.9448} \\
      SAM2  & 4.0844 & 1.8994 & 19.3595 & 3.2741 \\
      SAM3  & 3.9106 & \textbf{1.6924} & 18.3659 & 2.9196 \\
      \bottomrule
    \end{tabular}%
    \endgroup
    }
    \vspace{-10pt}
\end{wraptable}

%% file: Figures/sam_logit_binary_comparison.tex
\begin{wrapfigure}{r}{0.3\linewidth}
    \centering
    \vspace{-\baselineskip}
    \includegraphics[width=\linewidth]{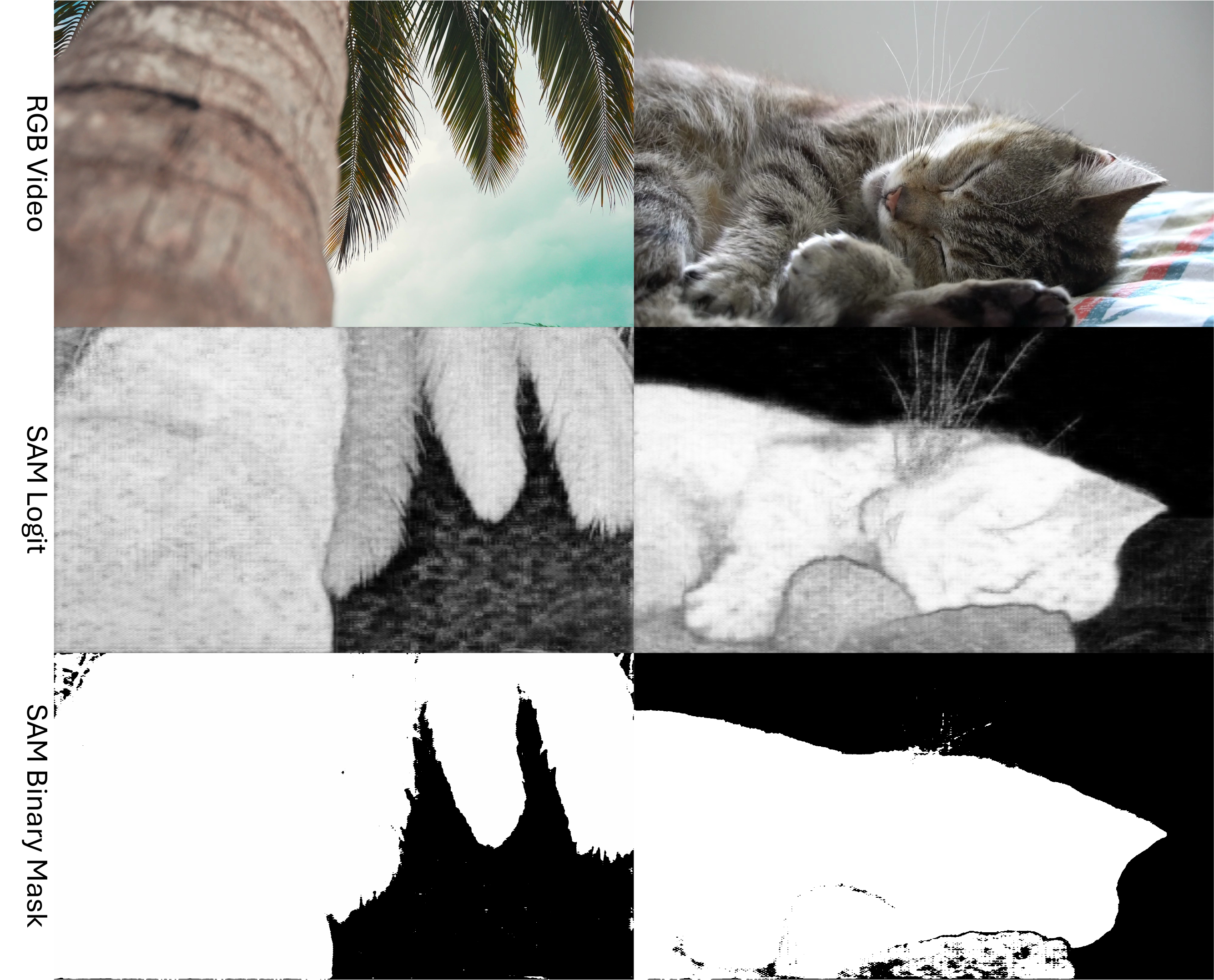}
    \vspace{-15pt}
    \caption{\textbf{SAM logit vs.\ Binary mask.}}
    \vspace{-10pt}
    \label{fig:sam_logit_binary_comparison}
\end{wrapfigure}

%% file: Tables/ablations.tex
\begin{table}[t]
  \centering
  \caption{\textbf{Ablation studies on video matting design choices.} 
    We evaluate the impact of our proposed components across different configurations. The ablations analyze (a) the advantage of incorporating SAM logits, and (b) channel-wise versus token-wise concatenation. All evaluations are performed on the \textbf{V-Him60 (Hard)} benchmark. The \textbf{best} results are highlighted.
  }
  \vspace{-3pt}
  \label{tab:ablations}

  \begin{subtable}[t]{0.48\linewidth}
    \centering
    \caption{\textbf{Advantage of SAM logit}}
    \label{tab:abl_samlogit}
    \resizebox{\linewidth}{!}{%
    \begin{tabular}{@{}l cccccc@{}}
      \toprule
      \textbf{Method} & MAD & MAD-T & MSE & GRAD & CONN & DTSSD \\
      \midrule
      with logit & \textbf{1.6272} & \textbf{40.8256} & \textbf{0.4247} & \textbf{2.9048} & \textbf{2.5843} & \textbf{1.9372} \\
      Ours & 1.6942 & 41.2279 & 0.4487 & 2.9196 & 2.7194 & 1.9796 \\
      \bottomrule
    \end{tabular}
    }
  \end{subtable}%
  \hfill
  \begin{subtable}[t]{0.48\linewidth}
    \centering
    \caption{\textbf{Channel vs. Token concat}}
    \label{tab:abl_input}
    \resizebox{\linewidth}{!}{%
    \begin{tabular}{@{}l cccccc@{}}
      \toprule
      \textbf{Method} & MAD & MAD-T & MSE & GRAD & CONN & DTSSD \\
      \midrule
      Channel & 2.2253 & 54.7369 & 0.6233 & 4.7119 & 3.8287 & 2.4183 \\
      Token & \textbf{1.6924} & \textbf{41.9167} & \textbf{0.4234} & \textbf{2.9813} & \textbf{2.7228} & \textbf{1.9758} \\
      \bottomrule
    \end{tabular}
    }
  \end{subtable}
  \vspace{-10pt}
\end{table}

%% file: Tables/abl_denoising_step.tex
\begin{wraptable}{r}{0.45\linewidth}
    \centering
    \vspace{-\baselineskip}
    \caption{\textbf{Ablation on denoising steps.} 
    }
    \label{tab:abl_denoising_steps}
    \vspace{-5pt}
    \resizebox{\linewidth}{!}{%
    \begin{tabular}{@{}l cccccc@{}}
      \toprule
      \textbf{Steps} & \textbf{MAD} $\downarrow$ & \textbf{MAD-T} $\downarrow$ & \textbf{MSE} $\downarrow$ & \textbf{GRAD} $\downarrow$ & \textbf{CONN} $\downarrow$ & \textbf{DTSSD} $\downarrow$ \\
      \midrule
      1  & \textbf{1.6680}  & \textbf{41.7354}  & \textbf{0.4027} & \textbf{2.9674} & \textbf{2.6769}  & 1.9546 \\
      2  & 1.7221  & 42.0193  & 0.4138 & 2.9708 & 2.7389  & \textbf{1.9479} \\
      3  & 2.7264  & 50.3829  & 0.5093 & 3.8081 & 3.3320  & 2.1253 \\
      5  & 10.7571 & 74.6095  & 2.0941 & 5.1494 & 17.5623 & 2.6155 \\
      10 & 16.6343 & 101.5099 & 3.9117 & 6.1175 & 31.1540 & 3.1936 \\
      \bottomrule
    \end{tabular}%
    }
    \vspace{-10pt}
\end{wraptable}

%% file: Tables/abl_chunksize.tex
\begin{wraptable}{r}{0.40\linewidth}
    \centering
    \vspace{-\baselineskip}
    \caption{\textbf{Ablation on chunk size.} 
    }
    \label{tab:abl_chunk_size}
    \vspace{-5pt}
    \resizebox{\linewidth}{!}{%
    \begin{tabular}{@{}l ccccc@{}}
      \toprule
      \textbf{Chunk} & \textbf{MAD} $\downarrow$ & \textbf{MAD-T} $\downarrow$ & \textbf{GRAD} $\downarrow$ & \textbf{CONN} $\downarrow$ & \textbf{DTSSD} $\downarrow$ \\
      \midrule
      3  & 2.1206 & 46.2131 & 3.5778 & 3.5776 & 2.2120 \\
      6  & \textbf{1.6942} & \textbf{41.2279} & \textbf{2.9196} & \textbf{2.7194} & \textbf{1.9796} \\
      12 & 1.7906 & 41.6290 & 3.0313 & 2.8858 & 2.0693 \\
      \bottomrule
    \end{tabular}%
    }
    \vspace{-10pt}
\end{wraptable}

%% file: Tables/abl_lora.tex
\begin{wraptable}{r}{0.40\linewidth}
    \centering
    \vspace{-\baselineskip}
    \caption{\textbf{Ablation on LoRA rank.} 
    }
    \label{tab:abl_lora_rank}
    \vspace{-5pt}
    \resizebox{\linewidth}{!}{%
    \begin{tabular}{@{}l cccccc@{}}
      \toprule
      \textbf{Rank} & \textbf{MAD} $\downarrow$ & \textbf{MAD-T} $\downarrow$ & \textbf{MSE} $\downarrow$ & \textbf{GRAD} $\downarrow$ & \textbf{CONN} $\downarrow$ & \textbf{DTSSD} $\downarrow$ \\
      \midrule
      64  & 2.4294 & 56.5292 & 0.8798 & 4.2180 & 4.2180 & 2.6284 \\
      128 & 2.3018 & 53.3100 & 0.8108 & 4.7875 & 3.9483 & 2.5732 \\
      512 & \textbf{1.6942} & \textbf{41.2279} & \textbf{0.4487} & \textbf{2.9196} & \textbf{2.7194} & \textbf{1.9796} \\
      \bottomrule
    \end{tabular}%
    }
    \vspace{-10pt}
\end{wraptable}

%% file: Section/5_Conclusion.tex
\section{Conclusion}
We introduce VideoMaMa++, a diffusion transformer model that leverages generative priors for robust video matting, overcoming data scarcity. VideoMaMa++ introduces preserve and refine tokens that enable temporally consistent matting on long videos, addressing a fundamental limitation of diffusion-based approaches confined to short clips. We demonstrate that VideoMaMa++ achieves state-of-the-art performance across diverse benchmarks while producing fine-grained, temporally consistent results on long in-the-wild videos. An annotation-free analysis further shows that our chunk-wise inference exhibits no measurable discontinuity at chunk boundaries, unlike VideoMaMa processing chunks independently. Nevertheless, VideoMaMa++ is more computationally demanding than feed-forward models and, like all mask-guided methods, depends on the upstream mask track; reducing the DiT inference cost and benchmarking temporal smoothness on real footage are promising future directions.
We believe VideoMaMa++ provides a scalable framework bridging the gap between short-clip generation and real-world long-video applications.

%% file: Section/X_Appendix.tex

\appendix

\section*{Appendix Overview}
This supplementary material provides additional details on implementation, evaluation, and qualitative results that could not be included in the main paper due to space constraints.


The document is organized as follows:

\begin{itemize}

    \item \textbf{Section~\ref{sec:additional_implementation}} provides additional implementation details, including the training configuration and the list of datasets used to train VideoMaMa++.

    \item \textbf{Section~\ref{sec:interactive_qual}} presents qualitative results for the interactive refinement experiment, illustrating how VideoMaMa++ progressively improves its own predictions across iterations through the preserve and refine token mechanism.

    \item \textbf{Section~\ref{sec:image_benchmark}} reports zero-shot evaluation of VideoMaMa++ on image matting benchmarks, demonstrating that our model generalizes effectively beyond video to single-image settings.

    \item \textbf{Section~\ref{sec:am2k}} discusses annotation noise in the AM-2K benchmark and explains why MAD-T provides a more reliable measure of matting quality than MAD on this dataset.

    \item \textbf{Section~\ref{sec:youtubematte_remask}} describes the guidance-mask generation protocol shared by all mask-guided methods and reports a single evaluation exception on the YouTubeMatte static\_0017 sequence, where SAM3~\cite{sam3} must be re-initialized at a later frame to capture an object that enters the scene partway through the video, together with results under first-frame initialization

    \item \textbf{Section~\ref{sec:app_seam}} introduces an annotation-free test for chunk-boundary discontinuities, which quantifies temporal consistency on in-the-wild videos without ground truth, and validates it on YouTubeMatte.

    \item \textbf{Section~\ref{sec:cost}} reports the computational cost of VideoMaMa++ and prior methods, including a per-stage runtime breakdown and end-to-end throughput with mask generation.
    
    \item \textbf{Section~\ref{sec:app_anchor}} details the reference-anchor experiments, including the anchor-refresh setting and the corruption protocol used to test robustness to a degraded anchor.

    \item \textbf{Section~\ref{sec:limitation}} discusses the limitations of this work: its dependence on the upstream mask track, and the absence of a video matting benchmark that can rigorously evaluate temporal smoothness on real-world footage.

    \item \textbf{Section~\ref{sec:societal_impact}} discusses the positive and negative societal impacts of mask-guided video matting.
\end{itemize}




\section{Additional Implementation Details}
\label{sec:additional_implementation}
\input{Tables/supplementary/data_use}

\paragraph{Training details.}
\label{sec:training_details}
We train VideoMaMa++ with an effective batch size of 16 and a learning rate of $1 \times 10^{-4}$ using the AdamW~\cite{adamw} optimizer. Rather than computing the loss in the latent space, we decode the predicted latents with the VAE decoder and supervise the resulting alpha mattes in pixel space. The training objective is a weighted sum of three terms:
\begin{equation}
    \mathcal{L} = \lambda_{\text{L1}}\,\mathcal{L}_{\text{L1}} + \lambda_{\text{grad}}\,\mathcal{L}_{\text{grad}} + \lambda_{\text{lap}}\,\mathcal{L}_{\text{lap}},
\end{equation}
where $\mathcal{L}_{\text{L1}}$ enforces pixel-wise accuracy, $\mathcal{L}_{\text{grad}}$ penalizes differences in spatial gradients to preserve edge sharpness, and $\mathcal{L}_{\text{lap}}$ is a Laplacian loss that emphasizes high-frequency structure such as fine boundaries and hair details. We set $\lambda_{\text{L1}} = 1.0$, $\lambda_{\text{grad}} = 5.0$, and $\lambda_{\text{lap}} = 2.0$. Training is conducted on 8 NVIDIA H100 GPUs for approximately 2.5 days. A detailed list of the training datasets is provided in Table~\ref{tab:existing_datasets}.



\section{Interactive Refinement Qualitative Results}
\label{sec:interactive_qual}
\input{Tables/ablation_iterative}
We provide qualitative results for the interactive refinement experiment in Figure~\ref{fig:interactive_qual}, complementing the quantitative results reported in Table~\ref{tab:abl_iterations}. Without modifying the input mask guidance, VideoMaMa++ progressively improves its own predictions across iterations by alternating preserve and refine assignments within each chunk. As shown in the figure, interactive refinement recovers regions missed in the initial prediction and sharpens fine-grained details along the foreground boundary, drawing on context from adjacent frames through the DiT's self-attention. This confirms that the gains in Table~\ref{tab:abl_iterations} stem from genuine cross-frame reasoning rather than simple smoothing, and demonstrates that the preserve and refine tokens enable the model to serve as its own corrector at inference time.

\section{Image benchmark}
\label{sec:image_benchmark}
\input{Tables/abl_p3m}
We further evaluate VideoMaMa++ in a zero-shot setting on P3M-10K~\cite{p3m} and AM-2K~\cite{am2k}, a portrait image matting benchmark and animal matting benchmark, to assess its generalization beyond video. Guide masks are obtained from SAM3~\cite{sam3} following the same procedure as in Section~\ref{sec:Quantitative}. Image matting benchmarks emphasize fine-grained detail recovery at high resolution, posing a stringent test for pixel-level accuracy. As reported in Table~\ref{tab:image_comparison}, VideoMaMa++ achieves state-of-the-art performance against prior portrait matting methods~\cite{sparsemat, lin2022robust, gvm}, despite not being specifically designed for portrait matting.
We note that AM-2K contains ground-truth annotation errors; our method shows inflated MAD scores on this benchmark despite producing visually superior mattes. Additional discussion on Section~\ref{sec:am2k}.

\input{Figures/Supplementary/AM-2k_justification}
\section{AM-2K Justification}
\label{sec:am2k}
We observe that the AM-2K~\cite{am2k} ground-truth annotations contain systematic errors that disproportionately penalize fine-detail predictions under the MAD metric. As illustrated in Figure~\ref{fig:am-2k_justification}, the AM-2K ground truth often lacks detail within foreground regions, despite the predictions correctly recovering them. Because MAD averages absolute differences over the entire image, these annotation gaps inflate the reported error even when the predicted matte is visually superior. MAD-T, by contrast, restricts the evaluation to the transition region defined by the ground-truth trimap, isolating the boundary detail where matting quality matters most. Under this more reliable metric, VideoMaMa++ achieves state-of-the-art performance against all comparison methods, indicating that the apparent MAD gap on AM-2K reflects annotation noise rather than a true performance deficit.

\section{Guidance Mask Protocol and YouTubeMatte SAM3 Mask}
\label{sec:youtubematte_remask}

\paragraph{Mask generation protocol.}
All mask-guided methods, including ours, receive the identical guidance mask track for every sequence, generated by Algorithm~\ref{alg:mask_protocol}. We binarize the ground-truth alpha matte of the initialization frame and prepend a duplicate of that frame to the video; for the default first-frame initialization, SAM3~\cite{sam3} thus processes the frame sequence $\{0, 0, 1, 2, \dots\}$. The binarized matte is given as a mask prompt on the prepended duplicate, SAM3 propagates it forward over the whole sequence, and we binarize the resulting per-frame mask logits while discarding the output on the duplicate. Every frame's guidance mask, including that of the initialization frame itself, is therefore a SAM3 prediction rather than the binarized ground truth. The initialization frame is the first frame for every sequence, with the single exception described below. No method therefore receives a privileged mask.

\begin{algorithm}[t]
\caption{Guidance mask generation, shared by all mask-guided methods.}
\label{alg:mask_protocol}
\begin{algorithmic}[1]
\Require Frames $\{I_t\}_{t=0}^{T-1}$, ground-truth mattes $\{\alpha^{\mathrm{gt}}_t\}_{t=0}^{T-1}$, initialization frame $s$ ($s{=}0$ except static\_0017)
\Ensure Binary mask track $\{M_t\}_{t=0}^{T-1}$
\State $m \gets \mathbf{1}\big[\alpha^{\mathrm{gt}}_s > \tau\big]$ \Comment{binarize the GT matte, $\tau = $ 127}
\State $\mathcal{J} \gets (I_s, I_0, I_1, \dots, I_{T-1})$ \Comment{prepend a duplicate; $s{=}0$ gives $\{0,0,1,2,\dots\}$}
\State Initialize the SAM3 video predictor on $\mathcal{J}$ with mask prompt $m$ at index $0$
\State Propagate forward over $\mathcal{J}$ to obtain mask logits $\ell_0, \ell_1, \dots, \ell_T$
\For{$t = 0$ \textbf{to} $T-1$}
    \State $M_t \gets \mathbf{1}\big[\ell_{t+1} > 0\big]$ \Comment{discard $\ell_0$, the output on the duplicate}
\EndFor
\State \Return $\{M_t\}_{t=0}^{T-1}$
\end{algorithmic}
\end{algorithm}

\input{Figures/Supple_YoutubeMatte_remask}
\paragraph{Re-initialized sequence.}
For all evaluations, the input masks are obtained automatically without human intervention, with one exception. The static\_0017 sequence from YouTubeMatte~\cite{matanyone} presents a granularity ambiguity at the first frame: a sheet of paper that becomes part of the foreground only enters the scene partway through the video, and is therefore not visible at the moment of mask initialization. As a result, when SAM3~\cite{sam3} is prompted on the first frame, the resulting mask track captures only the subject and cannot recover the paper once it appears, as shown in Figure~\ref{fig:sup_youtubematte_justification}. Because mask-guided matting models, including MaGGIe~\cite{maggie}, VideoMaMa~\cite{videomama}, and ours, faithfully follow the granularity defined by the input mask, the paper is consequently excluded from the predicted alpha matte. To resolve this without manual annotation, we initialize SAM3 on a later frame (frame 60) where both the subject and the paper are simultaneously visible, and propagate the prompt across the entire sequence. This re-initialized track is used for all mask-guided methods. We note that the MatAnyone~\cite{matanyone, matanyone2} series is not affected by this issue, as it does not rely on an external mask track. We report this exception in the interest of full transparency.

\begin{table}[t]
\centering
\caption{\textbf{Effect of re-initializing YouTubeMatte static\_0017.} ``Re-init'': SAM3 re-initialized at frame 60 for static\_0017, as used in Table~\ref{tab:quan}. ``First-frame'': SAM3 initialized on the first frame for every sequence. All metrics $\downarrow$.}
\label{tab:ytm_reinit}
\small
\begin{tabular*}{\textwidth}{@{\extracolsep{\fill}}llcccccc@{}}
\toprule
Metric & Init. & MGM & MGM-IW & MaGGIe & MaGGIe-FT & VideoMaMa & Ours \\
\midrule
\multirow{2}{*}{MAD}   & First-frame & 3.87 & 2.87 & 2.29 & 2.30 & 1.88 & \textbf{1.61} \\
                       & Re-init     & 3.59 & 2.41 & 1.67 & 1.52 & 1.41 & \textbf{1.15} \\
\midrule
\multirow{2}{*}{MSE}   & First-frame & 0.78 & 0.83 & 0.93 & 0.95 & 0.76 & \textbf{0.69} \\
                       & Re-init     & 0.53 & 0.38 & 0.50 & 0.37 & 0.30 & \textbf{0.23} \\
\midrule
\multirow{2}{*}{Grad}  & First-frame & 4.37 & 4.37 & 7.02 & 6.42 & 4.20 & \textbf{3.31} \\
                       & Re-init     & 4.27 & 4.25 & 5.11 & 5.36 & 4.08 & \textbf{3.27} \\
\midrule
\multirow{2}{*}{dtSSD} & First-frame & 1.61 & 1.40 & 1.64 & 1.58 & 1.33 & \textbf{1.20} \\
                       & Re-init     & 1.57 & 1.37 & 1.46 & 1.47 & 1.31 & \textbf{1.17} \\
\bottomrule
\end{tabular*}
\end{table}

\paragraph{Results under first-frame initialization.}
Table~\ref{tab:ytm_reinit} compares two settings on YouTubeMatte: the re-initialized track used in the main paper, and first-frame initialization for every sequence. All mask-guided methods degrade under first-frame initialization, since the sheet of paper is then missing from their guidance, and VideoMaMa++ remains the best method in both settings. The degradation itself reflects the behavior described above: the predicted matte follows the granularity of its guidance mask. Because a guidance mask is cheap to inspect and correct while a matte is not, we regard this faithful mask following as a desirable property for practical pipelines.

\section{Annotation-free Temporal Consistency}
\label{sec:app_seam}
Ground-truth alpha mattes are unavailable for long real-world footage, so reference-based temporal metrics such as dtSSD cannot be computed on in-the-wild videos (Section~\ref{sec:limitation}). Chunk-wise inference, however, has a characteristic failure mode that can be measured without annotation: a discontinuity where separately processed chunks meet. We therefore introduce a ground-truth-free test for such discontinuities.
 
\paragraph{Seam signal.}
Given a predicted matte sequence $\{\hat{\alpha}_t\}$, we compute the frame-to-frame change
\begin{equation}
    d_t = \mathcal{A}\big(\lvert \hat{\alpha}_{t+1} - \hat{\alpha}_t \rvert\big),
    \qquad
    \tilde{d}_t = d_t - \operatorname{med}\big(d_{t-w}, \dots, d_{t+w}\big),
    \label{eq:seam_signal}
\end{equation}
where $\mathcal{A}$ aggregates over pixels, either as the mean over foreground pixels (fg-mean) or as the maximum (MAX).
Subtracting a running median removes change caused by genuine motion, so $\tilde{d}_t$ retains only spikes above the local neighborhood.
 
\paragraph{Chunk cadence.}
To remove any confound from the two models tiling time differently, we run VideoMaMa~\cite{videomama} and VideoMaMa++ with the same chunk cadence: a 6-frame first chunk followed by 4 new frames per chunk. With frames indexed from 0 and transition $t$ denoting the change from frame $t$ to frame $t{+}1$, seams fall at transitions $\mathcal{S} = \{5, 9, 13, \dots\}$ for both models. A 100-frame YouTubeMatte clip thus has 99 transitions, of which 24 are seams ($5, 9, \dots, 97$) and 75 are interior. VideoMaMa processes each chunk without overlap, whereas VideoMaMa++ additionally conditions each chunk on its preserve-token frames.

\paragraph{Statistics.}
We test whether the spikes in $\tilde{d}_t$ recur at the chunk period with two statistics. First, we compute the autocorrelation of $\tilde{d}_t$ per clip at the seam period (lag 4) and at twice the seam period (lag 8), and average it over clips.  Second, we call a clip \emph{seam-elevated} if the mean of $\tilde{d}_t$ over $t \in \mathcal{S}$ exceeds its mean over interior transitions, and refer to the gap between the two as its \emph{seam contrast}. Under the null hypothesis of no chunk-boundary discontinuity, a clip is seam-elevated with probability $1/2$, which the ground-truth control confirms empirically (Table~\ref{tab:seam}). For the seam-elevated counts, we report p-values from an exact one-sided sign test against this chance level.

\begin{table}[t]
\centering
\caption{\textbf{Annotation-free seam test.} Both models run at identical chunk seams. \emph{Left:} ground-truth validation on YouTubeMatte (64 clips; 24 seam vs.\ 75 interior transitions per clip). Entries count clips whose mean seam transition exceeds their mean interior transition (chance: 32/64). \emph{Right:} 16 in-the-wild clips (fg-mean), for which no ground truth exists. Higher autocorrelation at the seam lag and more seam-elevated clips indicate a stronger chunk-boundary discontinuity. p-values for seam-elevated clips are from an exact one-sided sign test against chance.}
\label{tab:seam}
\small
\begin{tabular*}{\textwidth}{@{\extracolsep{\fill}}lccccc@{}}
\toprule
 & \multicolumn{2}{c}{YouTubeMatte} & \multicolumn{3}{c}{In-the-wild (no GT)} \\
\cmidrule(lr){2-3} \cmidrule(l){4-6}
 & \multicolumn{2}{c}{Seam-elevated clips} & \multicolumn{2}{c}{Autocorr.} & Seam-elevated \\
\cmidrule(lr){2-3} \cmidrule(lr){4-5}
Model & MAX & fg-mean & lag 4 & lag 8 & clips \\
\midrule
VideoMaMa~\cite{videomama} & 62/64 & 63/64 & $+0.35$ & $+0.34$ & 14/16 \\
 & {\scriptsize $p{=}1.1{\times}10^{-16}$} & {\scriptsize $p{=}3.5{\times}10^{-18}$} & {\scriptsize $p{=}5{\times}10^{-3}$} & {\scriptsize $p{=}6{\times}10^{-3}$} & {\scriptsize $p{=}2.1{\times}10^{-3}$} \\[1pt]
GT (no chunking) & 34/64 & 32/64 & -- & -- & -- \\
 & {\scriptsize $p{=}0.35$} & {\scriptsize $p{=}0.55$} & & & \\[1pt]
VideoMaMa++ (Ours) & 18/64 & 18/64 & $+0.03$ & $+0.11$ & 7/16 \\
 & {\scriptsize $p{>}0.99$} & {\scriptsize $p{>}0.99$} & {\scriptsize $p{=}0.66$} & {\scriptsize $p{=}0.12$} & {\scriptsize $p{=}0.77$} \\
\bottomrule
\end{tabular*}
\end{table}
 
\paragraph{Validation on YouTubeMatte.}
We first validate the test on YouTubeMatte, where ground truth is available (Table~\ref{tab:seam}, left). VideoMaMa's seam transitions exceed its interior transitions in nearly all clips under both summaries. As a control, we apply the identical test to the ground-truth mattes at the same transition positions. Since the ground truth involves no chunking, any elevation found there would indicate that the test responds to content rather than to chunk boundaries; it detects nothing, with counts at the chance level. VideoMaMa++ shows no seam elevation: its seams exceed its interior transitions in fewer clips than expected by chance under both summaries.
Per clip, VideoMaMa's seam contrast exceeds that of VideoMaMa++ in 62--64 of 64 clips (sign test, $p < 10^{-15}$).
 
\paragraph{In-the-wild results.}
On the 16 in-the-wild videos (Table~\ref{tab:seam}, right), the detrended signal of VideoMaMa autocorrelates at the chunk period (lags 4 and 8), and its seam transitions exceed its interior transitions in most clips. At identical seams, VideoMaMa++ shows none of this structure: neither autocorrelation is significant, and the number of seam-elevated clips is consistent with chance. These results quantitatively corroborate Figure~\ref{fig:main_qual_long}: at matched boundaries, VideoMaMa exhibits a chunk-boundary discontinuity that VideoMaMa++ does not.

\paragraph{Robustness and scope.}
These conclusions are unchanged across alternative difference statistics and detrending windows of 3 to 11 frames. The test targets only discontinuities at chunk boundaries: it measures neither per-frame accuracy nor flicker within a chunk, and it applies only to methods that process video in chunks, which is why the comparison is restricted to VideoMaMa.

\section{Computational Cost}
\label{sec:cost}
\begin{table}[t]
\centering
\caption{\textbf{Computational cost} on a single H100 at resolutions $256^2$ to $2048^2$ (1K\,=\,$1024^2$, 2K\,=\,$2048^2$), excluding mask generation for all methods. For VideoMaMa++, the cost of each 6-frame window is amortized over its 4 new frames, so FPS accounts for the recomputed overlap frames. MatAnyone/MatAnyone~2 and MGM/MGM-ITW share architectures and are listed once.
}
\label{tab:efficiency}
\footnotesize
\setlength{\tabcolsep}{2pt}
\begin{tabular*}{\textwidth}{@{\extracolsep{\fill}}l*{15}{c}@{}}
\toprule
 & \multicolumn{5}{c}{FPS $\uparrow$} & \multicolumn{5}{c}{Peak memory (GB) $\downarrow$} & \multicolumn{5}{c}{TFLOPs $\downarrow$} \\
\cmidrule(lr){2-6} \cmidrule(lr){7-11} \cmidrule(l){12-16}
Method & 256 & 512 & 768 & 1K & 2K & 256 & 512 & 768 & 1K & 2K & 256 & 512 & 768 & 1K & 2K \\
\midrule
MGMatting   & 66.71 & 66.23 & 63.56 & 60.44 & 18.05 & 0.58 & 0.84 & 1.37 & 1.98 & 5.41 & 0.036 & 0.115 & 0.240 & 0.411 & 1.55 \\
MaGGIe      & 23.2 & 19.3 & 20.0 & 15.0 & 7.5 & 0.52 & 1.18 & 2.52 & 4.88 & 16.01 & 0.021 & 0.083 & 0.187 & 0.332 & 1.33 \\
MatAnyone   & 43.6 & 42.0 & 38.9 & 34.6 & 7.5 & 0.33 & 0.67 & 1.32 & 3.43 & 22.40 & 0.030 & 0.124 & 0.289 & 0.542 & 2.94 \\
VideoMaMa   & 27.56 & 13.64 & 6.06 & 3.41 & 0.71 & 4.75 & 5.36 & 6.41 & 7.88 & 17.05 & 1.60 & 6.55 & 15.4 & 28.9 & 159 \\
VideoMaMa++ & 16.4 & 6.9 & 2.6 & 1.17 & 0.11 & 3.73 & 5.17 & 7.86 & 11.33 & 35.59 & 4.25 & 24.8 & 85.4 & 225 & 2920 \\
\bottomrule
\end{tabular*}
\end{table}

\begin{table}[t]
\centering
\caption{\textbf{Per-output-frame cost of VideoMaMa++} (ms) on a single H100, including SAM3 mask propagation. Matting-stage costs are measured per 6-frame window and divided by its 4 new frames. E2E FPS $= 1000\,/$ (SAM3 + matting total).}
\label{tab:runtime_breakdown}
\small
\begin{tabular*}{\textwidth}{@{\extracolsep{\fill}}lccccccc@{}}
\toprule
Res. & SAM3 & VAE Enc. & DiT (1-step) & VAE Dec. & Matting total & Matting FPS & E2E FPS \\
\midrule
256$^2$  & 45.4 & 24.9  & 20.1   & 15.9  & 60.9   & 16.4 & 9.4  \\
512$^2$  & 45.4 & 37.9  & 79.2   & 27.5  & 144.6  & 6.9  & 5.3  \\
768$^2$  & 45.4 & 78.3  & 254.3  & 55.8  & 388.4  & 2.6  & 2.3  \\
1024$^2$ & 45.4 & 139.0 & 616.1  & 97.7  & 852.8  & 1.17 & 1.11 \\
2048$^2$ & 45.4 & 681.0 & 7602.1 & 448.6 & 8731.7 & 0.11 & 0.11 \\
\bottomrule
\end{tabular*}
\end{table}

Table~\ref{tab:efficiency} reports throughput, peak memory, and compute on a single H100 from $256^2$ to $2048^2$. VideoMaMa++ is the most expensive of the compared methods: at $1024^2$ it runs at 1.17 FPS with 11.3\,GB of peak memory, compared to 3.41 FPS and 7.9\,GB for VideoMaMa, while the feed-forward baselines run at 15--60 FPS. Table~\ref{tab:runtime_breakdown} breaks down this cost by stage. At $1024^2$, the single DiT forward pass accounts for 616\,ms of the 853\,ms per output frame, with the VAE encoder and decoder taking 139\,ms and 98\,ms. SAM3 mask propagation runs once per frame at a fixed resolution and adds 45.4\,ms per frame regardless of the matting resolution, giving an end-to-end rate of 1.11 FPS at $1024^2$ and 9.4 FPS at $256^2$. Since the DiT pass dominates, single-step inference (Table~\ref{tab:abl_denoising_steps}) is essential for keeping this cost practical, as each additional denoising step would add a full DiT evaluation per window.

\section{Reference Anchor Analysis Details}
\label{sec:app_anchor}
This section details the reference-anchor experiments of Table~\ref{tab:anchor}, all conducted on V-HIM60~\cite{maggie}.

\paragraph{Anchor choice and refresh.}
The first frame may not be representative of later frames. We therefore replace it as the anchor with a random frame or the last frame of the first chunk, which leaves performance nearly unchanged across all metrics. Because the foreground may also change in appearance over long sequences, we further re-anchor to a recent frame every 3 chunks. V-HIM60 sequences contain substantial pose, scale, and appearance variation, yet refreshing yields nearly identical quality to the default setting, so a first-frame anchor does not become stale on this benchmark. Refreshing nevertheless remains available at negligible cost for footage whose appearance changes more drastically.
 
\paragraph{Corruption protocol.}
We corrupt the input of the first frame, which serves as the anchor, rather than the propagated matte. Because the anchor is the model's own first-frame prediction, a corrupted input yields a degraded anchor matte that is propagated to all later chunks. We consider three corruptions:
\begin{itemize}
    \item \textbf{Poor guidance mask.} We erode (under-segmentation, E) or dilate (over-segmentation, D) the first-frame guidance mask using grayscale morphology with a square kernel of $k \in \{5, 21\}$\,px.
    \item \textbf{Motion blur.} We apply a horizontal motion-blur kernel of $k \in \{7, 15, 25\}$\,px to the first-frame RGB (B), so the model must matte a motion-blurred appearance.
    \item \textbf{Occlusion.} We composite a real animal foreground from AM-2K~\cite{am2k} over the subject using its ground-truth matte, enlarging it until at least 99\% of the subject is covered, and zero the occluded region of the guidance mask so that it remains consistent with the now-invisible subject (Occ.).
\end{itemize}
We exclude the corrupted first frame from evaluation, so the metrics measure how the degraded anchor propagates to the rest of the video. This matters most under occlusion, where the input RGB and the ground-truth matte of the first frame disagree. It is also why the clean baseline in Table~\ref{tab:anchor}b differs slightly from the main results: the only difference is the excluded clean first frame.
 
\paragraph{Results.}
Across all corruptions, performance degrades only marginally relative to the clean baseline. Full occlusion is the most severe case, followed by motion blur and poor guidance masks, and larger kernels generally lead to slightly larger degradation. Even with a fully occluded anchor, the model remains close to its clean-anchor performance, indicating that a degraded anchor has only a limited effect on the rest of the video.

\section{Limitation}
\label{sec:limitation}
\paragraph{Dependence on the mask track.}
VideoMaMa++ deliberately separates tracking, delegated to a mask generator such as SAM3~\cite{sam3}, from matting, so its robustness is bounded by the quality of the mask track. Errors confined to individual frames of a chunk, such as missing thin structures or drift under motion blur, can be corrected through cross-frame attention to neighboring frames or through interactive refinement (Section~\ref{sec:interactive_qual}). However, if the tracker locks onto the wrong object for an entire sequence, the matte follows that wrong track. This limitation is shared by all mask-guided matting methods.

\paragraph{Evaluating temporal smoothness on real video.}
A key limitation of this work, and of video matting research more broadly, is the absence of a benchmark that can rigorously evaluate temporal smoothness on real video. Existing benchmarks~\cite{video240k, matanyone, videomama} are constructed by compositing per-frame foregrounds onto random backgrounds, which provides accurate per-frame ground truth but does not capture the temporal dynamics of naturally captured footage.
As a result, temporal consistency on real-world videos is still largely assessed qualitatively. The annotation-free seam test of Section~\ref{sec:app_seam} quantifies one specific failure mode, discontinuities at chunk boundaries, without ground truth. However, it measures neither per-frame accuracy nor flicker within a chunk, and it applies only to methods that process video in chunks, so it cannot compare the overall temporal behavior of arbitrary methods.
Developing a benchmark with reliable ground-truth temporal smoothness, ideally captured rather than composited, remains an open problem for the community.

\section{Societal Impact}
\label{sec:societal_impact}
\paragraph{Positive Societal Impact.}
Our mask-guided video matting framework can substantially benefit creative industries, education, and communication. By reducing the manual effort required for high-quality video matting, the method lowers the barrier to professional-grade video production, enabling independent creators, educators, and small studios to produce polished content without specialized expertise or costly proprietary software. In film and visual effects, it can accelerate labor-intensive rotoscoping workflows, allowing artists to focus on creative decisions rather than frame-by-frame manual labeling. The technology also improves the quality of background replacement in video conferencing and virtual classrooms, supporting remote collaboration and accessibility-oriented applications such as isolating speakers or interpreters in educational content.

\paragraph{Negative Societal Impact.}
Despite these benefits, high-quality video matting can be misused. As a core component in video manipulation pipelines, improved matting may facilitate the creation of more convincing deepfakes and manipulated media, contributing to misinformation and the erosion of public trust in video evidence.


%% file: Tables/supplementary/data_use.tex
\begin{wraptable}{r}{0.5\textwidth}
\centering
\vspace{-10pt}
\caption{\textbf{Datasets used to train VideoMaMa and SAM2-Matte.}}
\label{tab:existing_datasets}
\vspace{-5pt}
\resizebox{\linewidth}{!}{%
  \begin{tabular}{@{}ll@{}}
  \toprule
  \textbf{Category} & \textbf{Datasets} \\
  \midrule
  Image       & DVM~\cite{DVM}, AM-2k~\cite{am2k}, DAViD~\cite{DAViD}, P3M-10k~\cite{p3m}, RefMatte~\cite{refmatte} \\
  Video       & CRGNN~\cite{crgnn}, VideoMatte240K~\cite{video240k}, DVM~\cite{DVM} \\
  Background & VideoMatting108~\cite{video108}, DVM~\cite{DVM} \\
  \bottomrule
  \end{tabular}%
}
\vspace{-10pt}
\end{wraptable}

%% file: Tables/ablation_iterative.tex
\begin{wraptable}{r}{0.45\linewidth} 
    \centering
    \vspace{-5pt}
    \caption{\textbf{Ablation on interactive refinement.} 
    }
    \label{tab:abl_iterations}
    \resizebox{\linewidth}{!}{%
    \begin{tabular}{@{}l cccccc@{}}
      \toprule
      \textbf{Iter.} & \textbf{MAD} $\downarrow$ & \textbf{MAD-T} $\downarrow$ & \textbf{GRAD} $\downarrow$ & \textbf{CONN} $\downarrow$ & \textbf{DTSSD} $\downarrow$ & \textbf{MSE} $\downarrow$ \\
      \midrule
      1 & 1.6924 & 41.9167 & 2.9813 & 2.7228 & 1.9758 & 0.4234 \\
      2 & 1.6688 & 41.8064 & 2.9677 & 2.6761 & 1.9589 & 0.3992 \\
      3 & \textbf{1.6634} & \textbf{41.7927} & \textbf{2.9633} & \textbf{2.6655} & \textbf{1.9545} & \textbf{0.3949} \\
      \bottomrule
    \end{tabular}%
    }
    \vspace{-10pt}
\end{wraptable}

%% file: Tables/abl_p3m.tex
\begin{wraptable}{r}{0.5\linewidth}
    \centering
    \vspace{-5pt}
    \caption{\textbf{Quantitative evaluation on P3M-500 and AM-2k datasets.} 
    }
    \label{tab:image_comparison}
    \resizebox{\linewidth}{!}{%
    \begin{tabular}{@{}l cccc ccc@{}}
      \toprule
      \multirow{2}{*}{\textbf{Method}} & \multicolumn{4}{c}{\textbf{P3M-500-NP}} & \multicolumn{3}{c}{\textbf{AM-2k}} \\
      \cmidrule(lr){2-5} \cmidrule(l){6-8}
      & \textbf{MAD} $\downarrow$ & \textbf{MSE} $\downarrow$ & \textbf{Grad} $\downarrow$ & \textbf{Conn} $\downarrow$ & \textbf{MAD} $\downarrow$ & \textbf{MAD-T} $\downarrow$ & \textbf{Grad} $\downarrow$ \\
      \midrule
      \textit{Mask free methods} \\
      RVM       & 10.89 & 6.57  & 14.97 & 18.81 & - & - & - \\
      SparseMat & 11.14 & 6.36  & 13.47 & 18.13 & - & - & - \\
      GVM       & 8.39  & 4.89  & 12.89 & 14.21 & - & - & - \\
      \midrule
      \textit{Mask guided methods} \\
      SAM3      & 5.22  & 2.31  & 21.79 & 8.37  & 6.8729 & 93.2540 & 27.8649 \\
      MaGGIe    & 4.68  & 1.54  & 12.07 & 7.78  & \textbf{6.7721} & 73.5696 & 11.2396 \\
      Ours      & \textbf{4.23}  & \textbf{1.23} & \textbf{10.00} & \textbf{6.90} & 9.1688 & \textbf{61.0324} & \textbf{8.3373} \\
      \bottomrule
    \end{tabular}%
    }
    \vspace{-10pt}
\end{wraptable}

%% file: Figures/Supplementary/AM-2k_justification.tex
\begin{figure}[htbp]
    \centering
    \includegraphics[width=\linewidth]{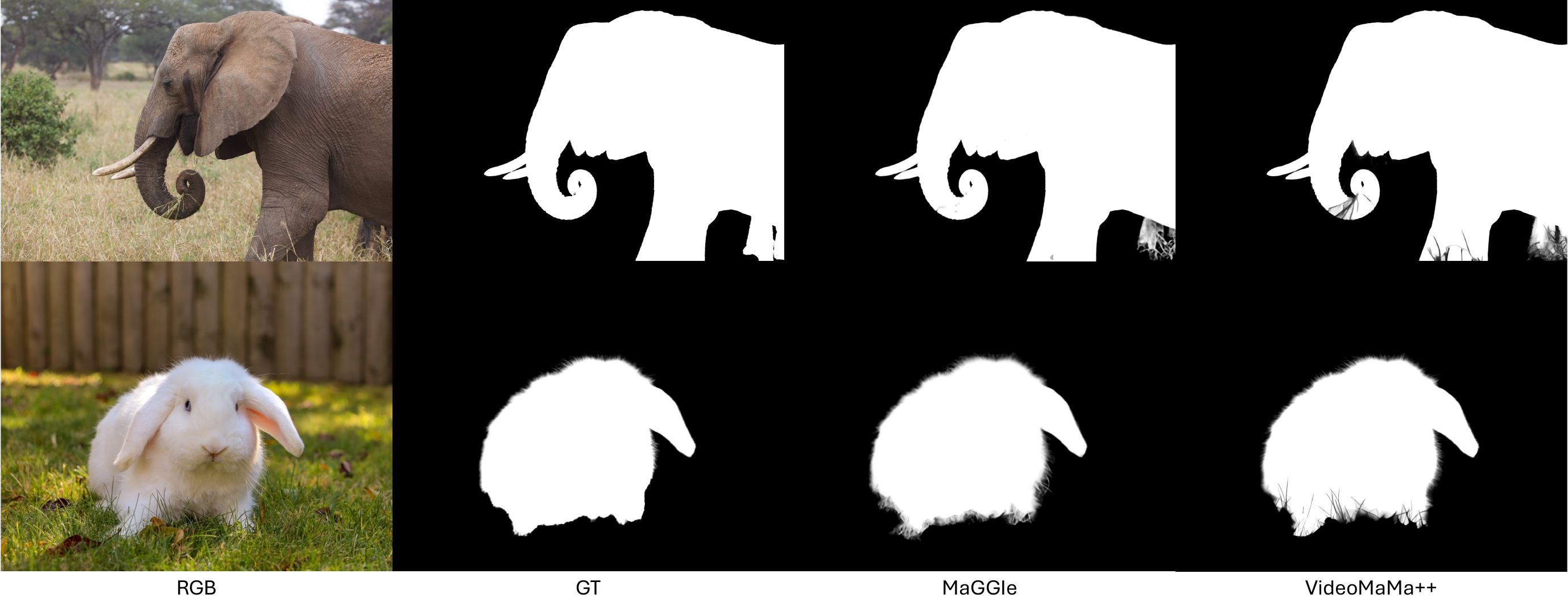}
    \caption{\textbf{Annotation noise in AM-2K~\cite{am2k}.} The ground-truth alpha mattes in AM-2K often miss fine-grained internal details, that VideoMaMa++ correctly recovers. Because MAD averages pixel-wise error over the entire image, these missing details penalize accurate predictions and inflate the reported error. 
    MAD-T, computed only within the transition region, isolates the matting quality that actually matters and avoids the bias introduced by such annotation gaps.
    }
    \label{fig:am-2k_justification}
\end{figure}

%% file: Figures/Supple_YoutubeMatte_remask.tex
\begin{figure}[htbp]
    \centering
    \includegraphics[width=\linewidth]{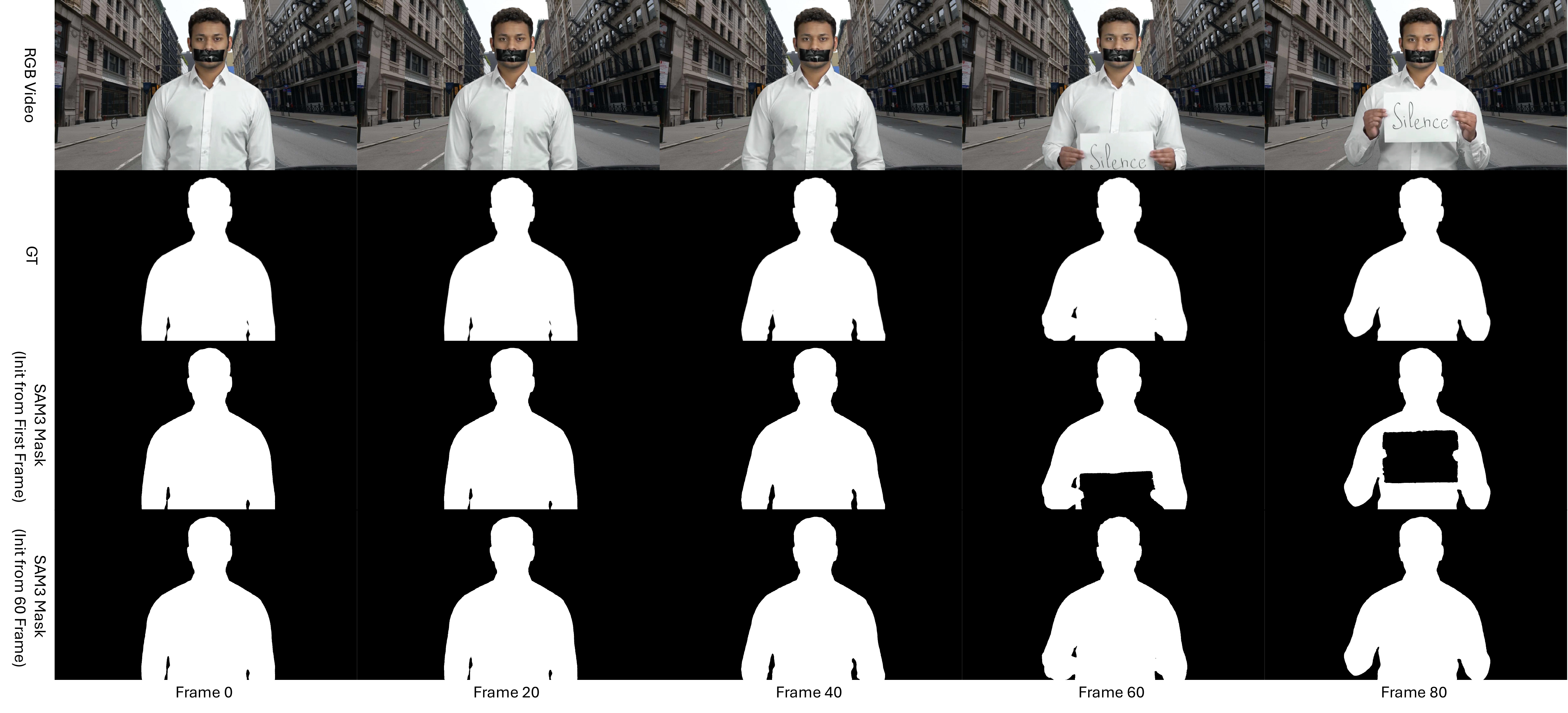}
    \caption{\textbf{Re-initialization of SAM3 on YouTubeMatte static\_0017.} Initializing SAM3~\cite{sam3} on the first frame fails to track a sheet of paper that the subject begins holding mid-sequence, since the object is not yet visible at initialization. Initializing on a later frame where both are simultaneously visible recovers the correct mask track.}
    \label{fig:sup_youtubematte_justification}
\end{figure}